\documentclass[journal,twoside]{IEEEtran}
\usepackage{amsmath,amsfonts}
\usepackage{algorithmic}
\usepackage{algorithm}
\usepackage{array}
\usepackage[caption=false,font=normalsize,labelfont=sf,textfont=sf]{subfig}
\usepackage{textcomp}
\usepackage{stfloats}
\usepackage{url}
\usepackage{verbatim}
\usepackage{graphicx}
\usepackage{balance}
\usepackage{cite}
\usepackage[table,xcdraw]{xcolor}
\usepackage{makecell}
\usepackage{booktabs} 
\usepackage{multirow}
\usepackage{hyperref}
\hypersetup{
	colorlinks=false,
}
\begin{document}

\title{Take an Irregular Route: Enhance the Decoder of Time-Series Forecasting Transformer}

\author{Li~Shen, Yuning~Wei, Yangzhu~Wang and~Huaxin~Qiu
\thanks{Manuscript received xxxx; revised xxxx. \textit{(Corresponding author: Li~Shen)}}
\thanks{\IEEEcompsocthanksitem Li Shen, Yuning Wei, Yangzhu Wang and Huaxin Qiu are with Beihang University, Beijing, China. (email: shenli@buaa.edu.cn; yuning@buaa.edu.cn; wangyangzhu@buaa.edu.cn; qiuhuaxin@buaa.edu.cn)}
}

\markboth{IEEE Internet OF Things Journal, }%
{Shen \MakeLowercase{\textit{et al.}}: Take an Irregular Route: Enhance the Decoder of Time-Series Forecasting Transformer}

\maketitle
\begin{abstract}
With the development of Internet of Things (IoT) systems, precise long-term forecasting method is requisite for decision makers to evaluate current statuses and formulate future policies. Currently, Transformer and MLP are two paradigms for deep time-series forecasting and the former one is more prevailing in virtue of its exquisite attention mechanism and encoder-decoder architecture. However, data scientists seem to be more willing to dive into the research of encoder, leaving decoder unconcerned. Some researchers even adopt linear projections in lieu of the decoder to reduce the complexity. We argue that both extracting the features of input sequence and seeking the relations of input and prediction sequence, which are respective functions of encoder and decoder, are of paramount significance. Motivated from the success of FPN in CV field, we propose FPPformer to utilize bottom-up and top-down architectures respectively in encoder and decoder to build the full and rational hierarchy. The cutting-edge patch-wise attention is exploited and further developed with the combination, whose format is also different in encoder and decoder, of revamped element-wise attention in this work. Extensive experiments with six state-of-the-art baselines on twelve benchmarks verify the promising performances of FPPformer and the importance of elaborately devising decoder in time-series forecasting Transformer. The source code is released in \url{https://github.com/OrigamiSL/FPPformer}. 
\end{abstract}


\section{Introduction}
\subsection{Background}
\IEEEPARstart{T}{he} advent of Big Data era has brought immense volume and variety of data in the 21st century, especially in Internet of Things (IoT) systems with tons of sensors \cite{IoT}. Consequently, it necessitates long-term time-series forecasting methods with demanding accuracy and efficiency to assist decision makers and engineers in the appraisal of sensor statuses and future plans. Since traditional forecasting methods based on statistics \cite{box1974,box2015} are no longer sufficient for such sophisticated situations, more and more data scientists pay their attention to deep time-series forecasting \cite{suvery}. After decades of development and competition, Time-Series Forecasting MLP (TSFM) \cite{DLinear,FiLM,NHiTS} and Time-Series Forecasting Transformer (TSFT) \cite{Crossformer,scaleformer,Triformer,MICN} become the mainstream.\par
\subsection{Problems}
TSFM and TSFT have different pros and cons. TSFM is known for its parsimonious but efficient architecture so that forecasting models based on TSFM excel in resisting non-stationarity brought by distribution shifts \cite{RevIN} and concept drifts \cite{Concept_Drift}. Conversely, forecasting models based on TSFT own more complicated architecture and better capability of capturing long-term dependencies of time-series at the expense of being more vulnerable to over-fitting problem caused by non-stationarity \cite{DLinear}. Fortunately, pioneers have striven to get around plenty of problems of TSFT. Direct forecasting strategy \cite{direct} reduces the time complexity and alleviates the error accumulation problem \cite{Informer}. RevIN \cite{RevIN} solves the problem of distribution shifts among windows with distinct time spans. The channel-independent \cite{PatchTST} forecasting method renders TSFT refraining from extracting vague inter-relationships of different variables. Patch-wise attention mechanism \cite{PatchTST,Triformer} further attenuates the space complexity and brings the capability of local feature extraction to TSFT. Indeed, recent works have proven that TSFT models \cite{non-stationaryTransformer,scaleformer} can also be stable and robust in forecasting. Evidently, the majority of these enhancement focus on improving the encoder architecture and tackling input sequence features. It cannot be denied that they are very important, but not solely. The connections of input and prediction sequences, manifested by decoder in TSFT, are also of paramount significance, especially for pursuing precise forecasting in IoT. However, its significance is frequently omitted and itself is inadequately explored. Normally, the decoder architectures of existing TSFT models are simply duplicates of their encoder architectures, barring with little indispensable modifications, such like changing self-attentions to cross-attentions \cite{Informer,Crossformer}. Furthermore, some researchers have gone so far as to substitute decoder in TSFT with simple linear projection \cite{PatchTST,Pyraformer}, which is analogous to TSFM, for the sake of enhancing their efficiency. Now it is time to enhance the decoder of TSFT to fully develop its potential and push its forecasting performances to a new altitude.\par
\subsection{Contributions}
Different from existing TSFT models, we \textbf{\textit{F}}ully develop the tried-and-tested \textbf{\textit{P}}atch-wise attention mechanism and \textit{\textbf{P}}yramid architecture in \textit{both encoder and decoder} and thereby propose FPPformer. Alike FPN \cite{FPN} and PAN \cite{PAN} architectures which are prevalent in CV fields, FPPformer hierarchically extracts input sequence features \textit{from fine to coarse} and constructs prediction sequence \textit{from coarse to fine}. To strengthen the feature extraction capability of patch-wise attention, we further insert an element-wise attention block into each patch-wise attention block to \textit{extract fine-grained inner-relations of each patch in encoder and decoder with merely linear complexity}. A \textit{channel-independent and temporal-independent embedding method} is utilized to modify the size of feature maps in FPPformer to meet the needs of element-wise attention and patch-wise attention. Within each attention block in encoder, \textit{the diagonal line of query-key matching matrix is masked} to ensure the generality of features extracted from input sequences. Primary contributions of this work are {five} folds:
\begin{enumerate}
\item We propose a novel time-series forecasting Transformer, i.e., FPPformer, which uncommonly and efficaciously improves the decoder architecture of TSFT to break its fetters and excavate its potential.
{ \item We renovate the decoder architecture of TSFT and change it into top-down architecture for the sake of rationally constructing the prediction sequence in a hierarchical manner.
\item Motivated by a pioneer anomaly detection method, we propose diagonal-masked self-attention to mitigate the negative impacts of the outliers in input sequences.}
\item A new combination of element-wise attention and patch-wise attention is proposed by us {to compensate the weakness of conventional patch-attention in extracting the inner-features of each patch, with only additional linear complexity}.

\item Extensive experiments under diverse settings validate that FPPformer is capable of reaching state-of-the-art on twelve benchmarks with peerless accuracy and robustness.
\end{enumerate}
\section{Related Works}
The past few years have witnessed the development and the success of deep-learning based forecasting methods. Thanks to the help of neural network, the long-term multivariate forecasting is no more a dream so that even IoT systems with plenty of sensors and explosive data can be predicted \cite{IoT,IoT1,IoT2}. Researchers have developed deep forecasting methods built upon diverse networks and Transformer is a hot topic among corresponding literature.\par
\paragraph{Time-Series Forecasting Transformer} Traditionally, Time-Series Forecasting Transformer (TSFT) executes the forecasting via encoder-decoder architecture. The Transformer encoder is used to extract the features of input sequence, then the Transformer decoder is able to construct the prediction sequence by the extracted features {of encoder and the prediction sequence, which is initialized with a certain number since it is unknown at the beginning}. These two processes are completed predominantly by \textit{attention mechanism}, thereby researchers always keep an eye on it. LogSparse Transformer \cite{logtrans} and Informer \cite{Informer} discover the sparsity of query-key matching matrix and they force the elements of query to attach to the partial elements of key for the sake of reducing the complexity. Autoformer \cite{Autoformer}, FEDformer \cite{FEDformer} and ETSformer \cite{ETSformer} combine the TSFT with seasonal-trend decomposition and signal processing method, e.g., Fourier Analysis, in attention mechanism to enhance their interpretability. Patch-wise attention is more popular and proven to be more useful recently. TSFTs with patch-wise attention, including Triformer \cite{Triformer}, Crossformer \cite{Crossformer} and PatchTST \cite{PatchTST}, achieve more promising performances than preceding models. However, whichever TSFT always emphasizes that the modified architecture is intended for \textit{more efficient or effective feature extraction for input sequence}. Hardly ever can statements \textit{involved with the profits of decoder} be found. Indeed, their decoders seem to play the role of requisite appendages in the entire Transformer architecture. Once some parts of encoders are changed by their proposed methods, mirrored changes are made to their decoders. Some researches \cite{PatchTST,Pyraformer} even abandon the decoder to circumvent these changes. Contrary to them, studying and figuring out the correct way of designing decoders in TSFT is exactly what this work is supposed to do.\par
\paragraph{Other Miscellaneous Deep Forecasting Methods}
Barring TSFT, there are plenty of other types of deep forecasting methods. Forecasting methods based on RNN and CNN are feasible ones. Their respective representatives LSTNet \cite{LSTNet} and SCINet \cite{SCINet} both achieved shiny performances during their periods. However, compared with the foregoing two types of deep forecasting methods, Time-Series Forecasting MLP (TSFM) relatively receives more attentions. These forecasting networks are solely comprised of linear-projection layers, whereas they still achieve promising performances. Due to their simple architectures, it is convenient for them to combine with statistics models for the objective of improving their interpretability and forecasting capability. NBEATS \cite{NBEATS} and DLinear \cite{DLinear} adopt seasonal-trend decomposition methods in their networks more concisely than FEDformer \cite{FEDformer} but achieve better results in general. C. Challu et al \cite{NHiTS} further presented N-HiTS that employs sampling and interpolation strategies on the basis of NBEATS for more precise and hierarchical prediction. Reconstruction method motivated from Legendre Polynomials is taken into account by T. Zhou et al \cite{FiLM} to come out with FiLM. TSMixer proposed by V. Ekambaram et al \cite{TSMixer} considers the temporal patterns, cross-variate information and additional auxiliary information to render TSFM ready for more complicated forecasting cases. They are challenging competitors for TSFT and we chiefly compare FPPformer with other TSFTs and these TSFMs in forthcoming experiments.\par
\section{Preliminary}
\paragraph{Problem Statement} 
This work primarily concentrates on multivariate forecasting problem. As the term suggests, a multivariate forecasting problem is to predict a certain window $ \{x_{t_2:t_3}\}^{1:V} $ with time duration of ($ t_3 - t_2 $) and variable number of $ V $ with its anterior window $ \{x_{t_1:t_2}\}^{1:V} $. Each $ x_t^v \in \mathbb{R} $, where $ t \in[t_1:t_3) $ and $ v \in[1, V] $, denotes an element at timestamp $ t $ and stemming from variable $ v $. There are quite a few nomenclature style to name the dimension of $ t $ and $ v $. In this work, the dimension of $ t $ is termed the temporal dimension and the dimension of $ v $ is termed the variable dimension. Note that the word `channel-independent' mentioned above refers to the independence at the variable dimension. Moreover, we still discuss several univariate forecasting cases where $ V = 1 $ in our experiments since the variable treatment strategies of different forecasting methods can be very distinctive, making multivariate forecasting comparison, albeit prevailing, not persuasive enough.\par
\begin{figure}[!t]
\centering
\includegraphics[width=3.3in]{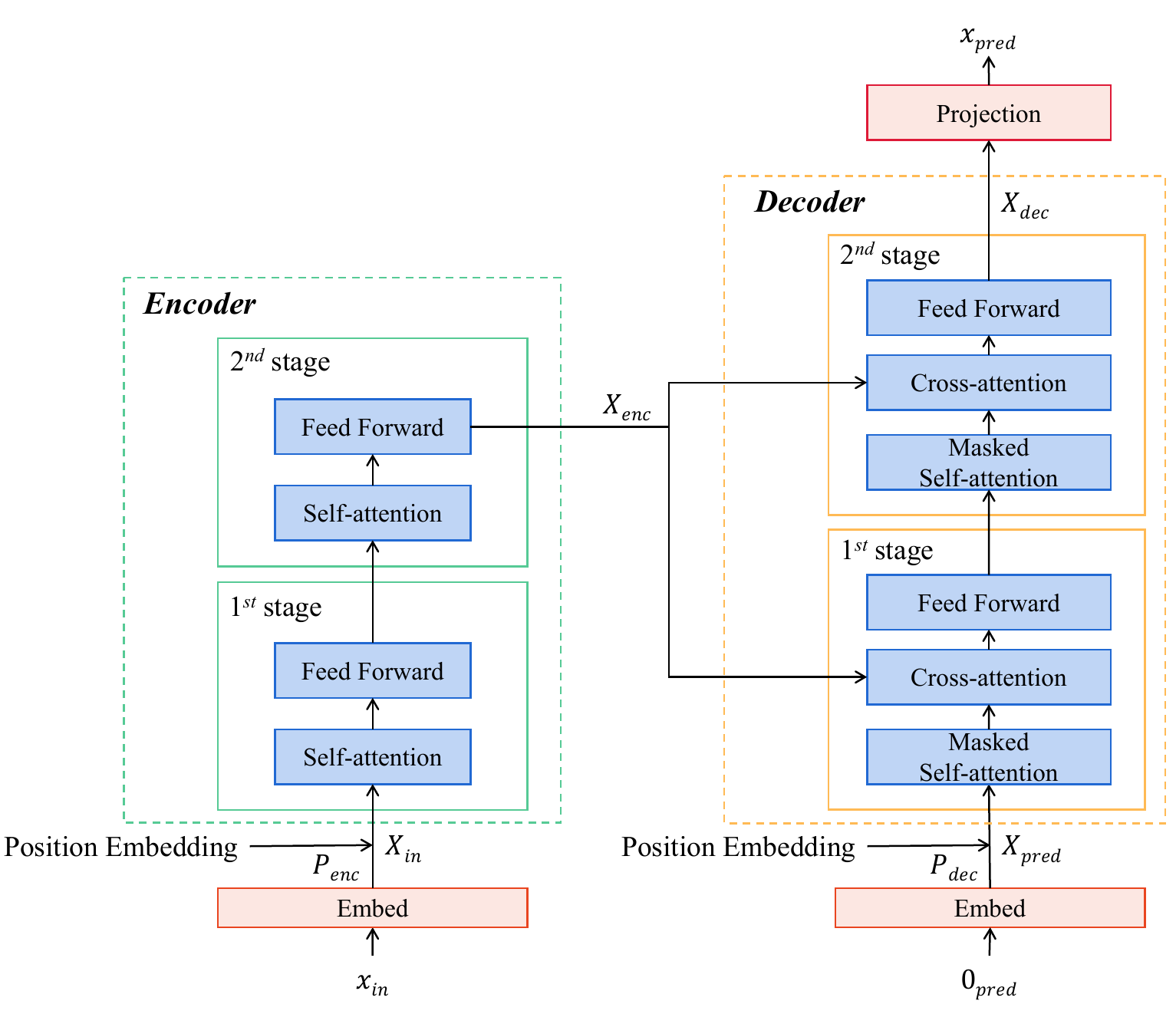}
\caption{A schematic of a vanilla TSFT with two-stage encoder (Green dashed box containing green solid boxes in the left) and two-stage decoder (Orange dashed box containing orange solid boxes in the right).}
\label{fig1}
\end{figure}

\paragraph{Vanilla TSFT Architecture} The architecture of vanilla TSFT is chiefly composed of \textit{an encoder, a decoder and a projection layer}. An example of a vanilla TSFT with 2-stage encoder and 2-stage decoder is sketched in Fig. \ref{fig1}. We can see that input sequence $ x_{in} $ passes through the \textit{encoder} embedding layer, then the superimposition of the embedded input sequence $ X_{in} $ and its position embedding $ P_{enc} $, which covers the input time span, is sent to the encoder. The encoder processes $ X_{in} + P_{enc} $ with $ M $ ($ M=2 $ in Fig. \ref{fig1}) stages, each consisting of a self-attention block and a feed forward layer, and the ultimate output feature map of the encoder is $ X_{enc} $. Typically, the canonical self-attention conducts scaled dot-products with the formula of (\ref{eq1}):
\begin{equation}
	\label{eq1}
	Attn(q, k, v) = \text{Softmax}(\frac{qk^\top}{\sqrt{D}}v)
\end{equation}

where $ q, k, v \in \mathbb{R}^{L\times D} $ are the linear projections of the identical sequence tensor, $ L $ is the number of tokens (or sequence tensor length) and $ D $ is the hidden dimension. Readers can refer to \cite{attention} to be familiar with attention mechanism and feed forward layer. Correspondingly, the \textit{decoder} receives both $ X_{enc} $ and the zero-initialized prediction sequence $ 0_{pred} $. $ 0_{pred} $ propagates through the decoder embedding layer to obtain $ X_{pred} $ and its position embedding $ P_{dec} $, which covers the prediction time span, is superimposed on it. Afterwards, they are sequentially sent into $ N $ ($ N=2 $ in Fig. \ref{fig1}) stages, each consisting of a masked self-attention block, a cross-attention block and a feed forward layer. Causality is essential as the prediction sequence is unknown, so that the masked self-attention, rather than normal self-attention, is utilized in decoder. The cross-attention block is intended to construct the prediction sequence via the encoder feature map $ X_{enc} $. Eventually, \textit{a projection layer} maps the output feature map of decoder $ X_{dec} $ to the prediction sequence $ x_{pred} $.\par
\paragraph{Employed Mechanisms} 
Barring our proposed methods, which will be introduced in the upcoming section, we also employ several advanced time-series forecasting mechanisms in FPPformer:
\begin{enumerate}
	\item \textit{Direct forecasting method} \cite{direct}, which is widely employed by recent deep forecasting method, performs the prediction of the entire sequence with only one forward process to alleviate the error accumulation.
	\item \textit{Channel-independent forecasting method}, which has been mentioned in the foregoing sections, treats the sequences of different variables as different instances. The sequences of different variables are parallel sent into the network without interfering with each other so that the network can seek shared characteristics of different variable sequences without imposing any inductive bias to the correlations of different variables.
	\item \textit{RevIN} \cite{RevIN}, which is devised for non-stationary time-series forecasting, normalizes each input sequence with its own statistics before sent into the network and restores the original statistics to the prediction sequence via the reverse instance normalization to handle the distribution shifts of real-world long time-series.
	\item \textit{Patch-wise attention} \cite{PatchTST}, which segments the sequence into patches of the same length, treats each single patch, rather than each single element, as a token in (\ref{eq1}) and treats the elements inside each patch or their latent representations as the hidden features of each token (patch) for better efficiency and generality.
\end{enumerate}

Note that the majority of recent deep forecasting models \cite{DLinear,NHiTS, PatchTST,Triformer,DishTS,FiLM,SCINet,Crossformer,TSMixer}, including those employed in our experiments, at least adopt two of above mechanisms so that they are not something that distinguish our methods from others.\par
\section{Methodology}
\subsection{Analysis of Decoder in TSFT}
\label{decoder}
Before commencing the introduction to our proposed FPPformer, we point out some deficiencies of current decoder architecture in TSFT to clarify the necessity and rationality of the decoder improvement in FPPformer.\par
We first discover the \textit{redundant self-attention problem in decoder}. To elaborate, we notice that the input to decoder in Fig. \ref{fig1} is a zero-initialized prediction sequence $ 0_{pred} $ owing to the unknown future. Consequently, the first (masked) self-attention is performed only on the position embedding of the prediction sequence. No matter it is fixed \cite{Informer} or learnable \cite{Crossformer}, it is completely independent of input sequence. \textit{Keep in mind that time-series forecasting is an auto-regressive problem.} It makes no sense to perform (masked) self-attention only on position embedding with the attempt of deducing some \textit{\textbf{groundless relations}}. Moreover, position embedding is always static after training while the input sequence can be dynamic and non-stationary, shattering the last hope that the position embedding can fit the statistics of time-series sequences due to some sort of assumptions with respect to homogeneity \cite{MetePFL,PromptCast}. Start token \cite{Informer} can be a solution, however short start token is not enough for long-term forecasting and longer start token brings about excessive complexity.\par
Besides, the connection of encoder and decoder is unitary, leading to the \textit{multi-scale insufficiency} problem. As shown in Fig. \ref{fig1}, the encoder with $ M $ stages can produce $ M $ feature maps of input sequence, however merely the last one is sent to decoder. This problem is more noteworthy when it comes to some modified TSFT with \textit{hierarchical architecture in encoder}. For instance, Informer \cite{Informer} employs convolution layers between every two stages in encoder and FEDformer \cite{FEDformer} keeps decomposing input sequences to acquire more precise seasonal features but \textbf{\textit{neither of them apply the same operations to decoder and only the feature map of the last stage in encoder is sent to decoder}}. Crossformer \cite{Crossformer} merges adjacent segments to obtain bigger patches in deeper stages in both encoder and decoder. However, just as what we claim in foregoing sections, the architecture of decoder is merely the replica of encoder with an additional cross-attention in Crossformer. By merging patches from small to big in decoder, Crossformer attempts to \textbf{\textit{construct the unknown prediction sequence from fine to coarse}}, whose irrationality is self-evident.\par
\subsection{Model Architecture}
\begin{figure}[!t]
\centering
\includegraphics[width=3.3in]{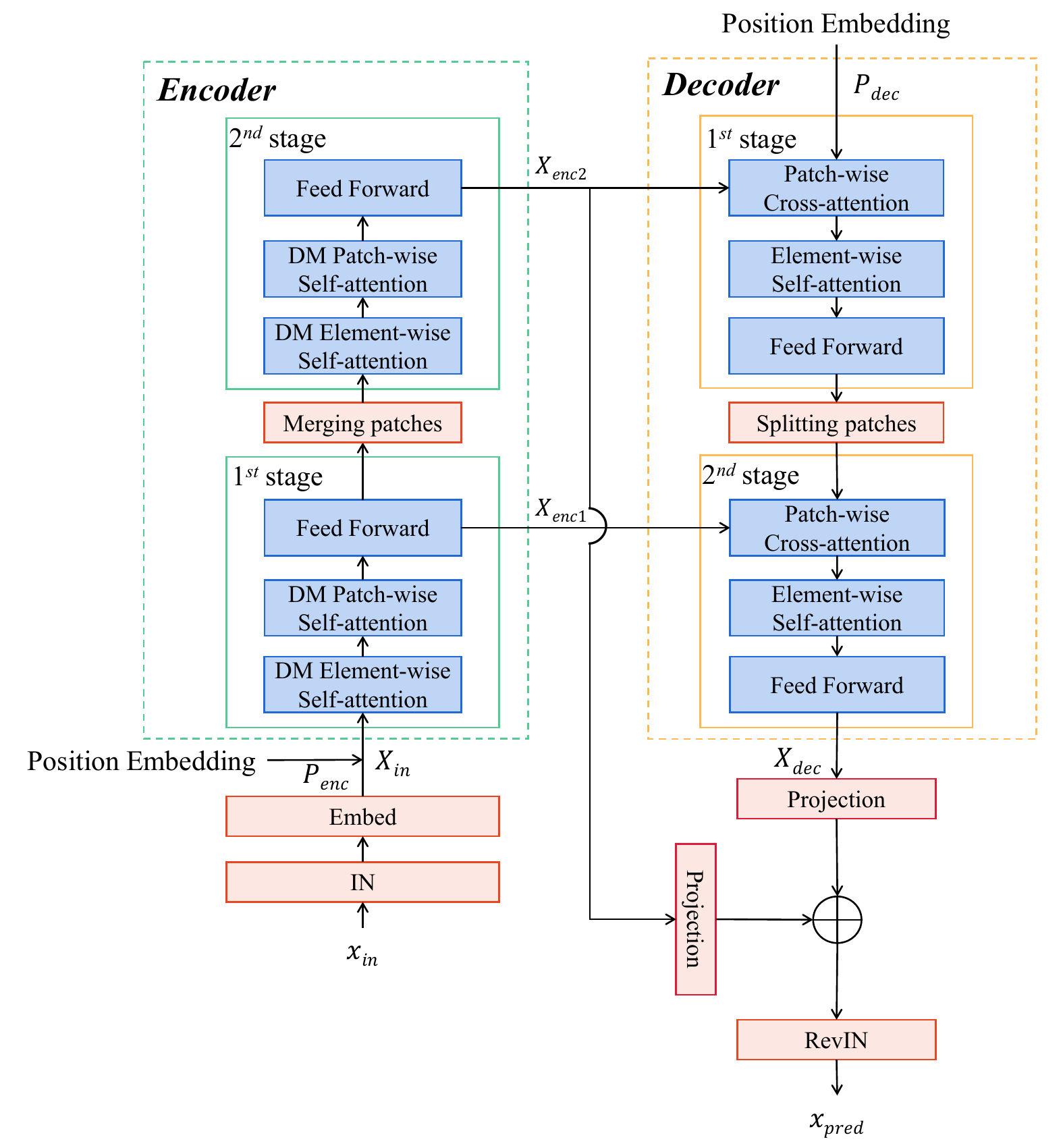}
\caption{An overview of FPPformer's hierarchical architecture with two-stage encoder and two-stage decoder. Different from the vanilla one in Fig. \ref{fig1}, the encoder owns bottom-up structure while the decoder owns top-down structure. Note that the direction of the propagation flow in decoder is opposite to that in Fig. \ref{fig1} to highlight the top-down structure. `DM' in the stages of encoder means `Diagonal-Masked'.}
\label{fig2}
\end{figure}
\begin{figure*}[]
	\centering
	\subfloat[FPPformer]{\includegraphics[width=3.3in]{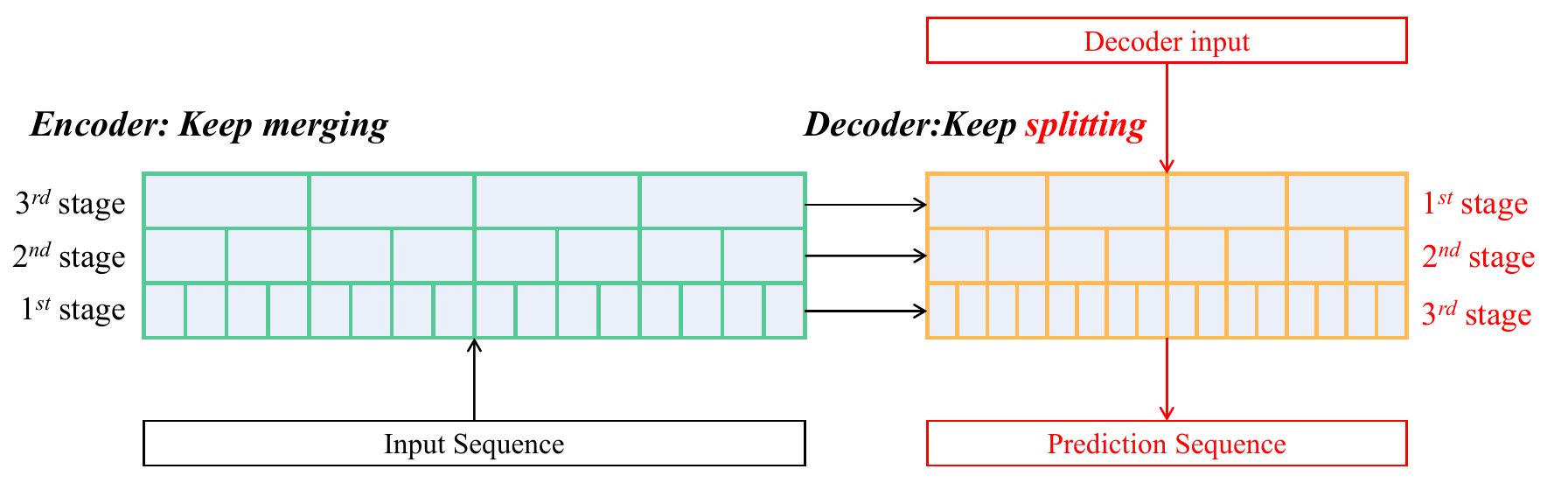}
	\label{fig3a}}
	\hfil
	\subfloat[Crossformer]{\includegraphics[width=3.3in]{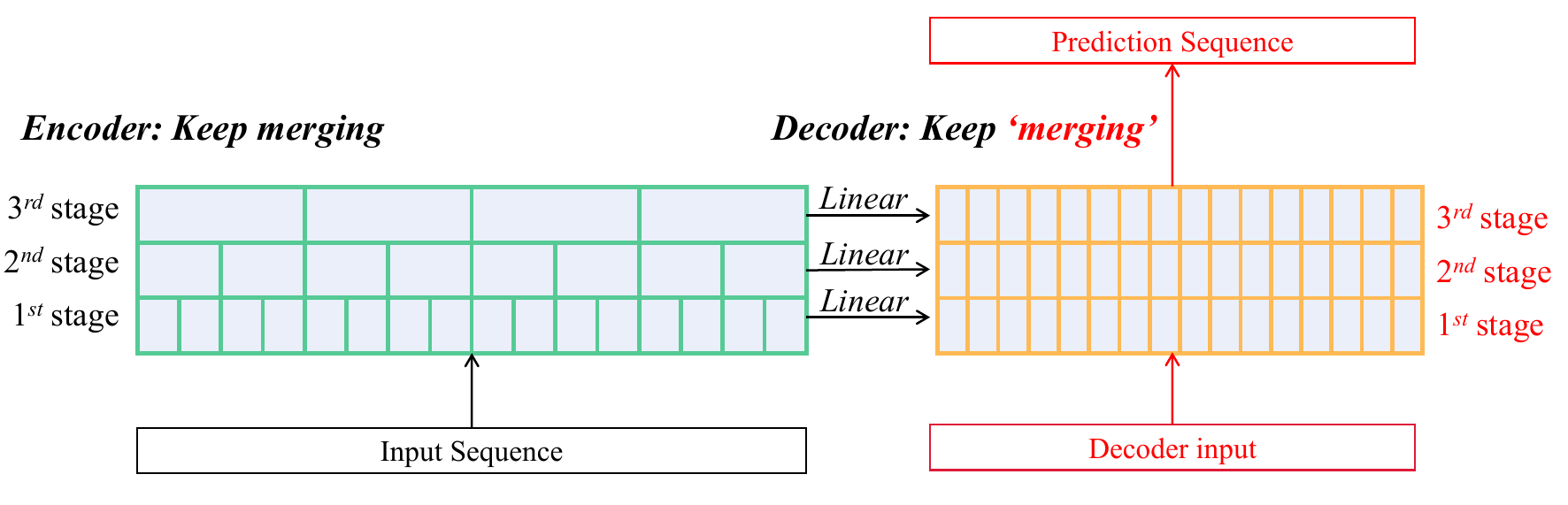}%
	\label{fig3b}}
	\caption{The comparison of hierarchical architecture of FPPformer (a) and Crossformer (b). The discrepancies are highlighted with red. Obviously, the decoder structure of Crossformer is nearly a duplicate of encoder's bottom-up structure whereas the decoder of FPPformer owns a different `top-down' structure.}
	\label{fig3}
\end{figure*}

The overview of our proposed FPPformer is illustrated in Fig. \ref{fig2} and its major enhancement on vanilla TSFT concentrates on addressing the preceding two problems of decoder. Comparing the schematics in Fig. \ref{fig1} and \ref{fig2}, the differences with respect to the overall architecture can be readily noticed. To handle the first redundant self-attention problem in decoder, we \textit{change the order of the self-attention and cross-attention} in decoder. Thereby, before embarking upon deducing any relations within unknown prediction sequence, the prediction sequence receives \textit{the auto-regressive parts from the deepest encoder feature map}, which serves as a better role for prediction sequence initialization before the first self-attention in decoder than simple zero-initialization with start token \cite{Informer}, randomly generated parameters \cite{Crossformer}, the trend decomposition of raw input sequence \cite{FEDformer}, and so forth. It is evident that the latter initialization formats of other TSFTs are either relatively simple or inefficient.\par
We employ the hierarchical pyramids both in encoder and decoder with lateral connections to tackle the second multi-scale insufficiency problem. As we adopt the patch-wise attention in FPPformer, the patches are merged before sent to the next stages in the bottom-up architecture of encoder and opposite operations, i.e., the splitting, are performed in the top-down architecture of decoder. The feature map of input sequence gets deeper and more coarse-grained within later stages, which is also the property shared by encoders of many TSFTs. However, things get different when we attempt to construct the prediction sequence from the position embedding and encoder feature maps. Recall how we decompose and reconstruct an arbitrary sequence from a certain multi-resolution analysis $ \{V_j\}_{j\in \mathbb{Z}} $ of $ L^2(\mathbb{R}) $ and wavelet spaces $ \{W_j\}_{j\in \mathbb{Z}} $ in wavelet theory \cite{Wavelet}, which owns a transcendent position in signal processing. When we decompose certain sequence $ f_{i+1}(t)\in V_{j+1} $, we decompose it into coarser spaces $ V_j $ and $ W_j $. Whereas the reconstruction is opposite, i.e., we recover the sequence in finer space $ f_{i+2}(t)\in V_{j+2} $ from $ V_{j+1} $ and $ W_{j+1} $. Omitting the existence of wavelet space $ W_j $, which contains the information of details or noises, we can find that \textit{the encoder and decoder processes separately correspond to the decomposition and (re)construction processes} in multi-resolution analysis. From another perspective, the unknown prediction sequence is initialized with zero or other parameters not pertaining to the ground truths at first. Thereby, when we strive to construct it from input sequence features, it is natural to commence with \textit{the most universal features to ensure the exactitude of general characteristics of prediction sequence features}, then we can prudently take steps to \textit{seek finer features of prediction sequence to avoid over-fitting}. The success of FPN \cite{FPN}, which also employs bottom-up and top-down architectures, in CV fields further confirms the preceding idea. Therefore, we keep splitting the patches in decoder and commence the hierarchical prediction sequence constructions with the feature maps from encoder, separately with identical resolutions, via lateral connections. The encoder in FPPformer presents a bottom-up architecture while the decoder presents a top-down architecture. Differences between the hierarchical design in Crossformer \cite{Crossformer}, whose decoder architecture is merely a replication of that encoder \footnote{In effect, the decoder of Crossformer even does not own a pyramid architecture. We use this statement since the hierarchical process of constructing the unknown prediction sequence is determined by how the model hierarchically uses the encoder features and for the more vivid comparison between FPPformer and Crossformer in Fig.\ref{fig3}.}. M. A. Shabani et al \cite{scaleformer} also notice the analogous thing but they neither expound the reasons of doing so nor they carry out any change to decoder.\par
\subsection{Combined Element-wise Attention and Patch-wise Attentions}
\label{patch}
The preceding two problems are shared by the majority of TSFTs and we would like to mention another specific problem of TSFT with patch-wise attention. We name this type of Transformer PTFST for brevity. Different from the element-wise attention of vanilla TSFT, which seeks the correlations of sequence elements, the patch-wise attention seeks the correlations of different patches or segments of sequences to improve the efficiency and reduce the risk of over-fitting. PatchTST \cite{PatchTST} and Crossformer \cite{Crossformer} are such PTSFTs and their experiments have proven the superiority of patch-wise attention. However, they neglect the inner-relations of the elements inside the patches or only employ simple linear projections to mix them up. Therefore, we make different changes in employed patch-wise attention in encoder and decoder for the sake of extracting more fine-grained features in encoder or pursuing finer prediction sequence construction in decoder.\par
\begin{figure}[!t]
\centering
\includegraphics[width=3.35in]{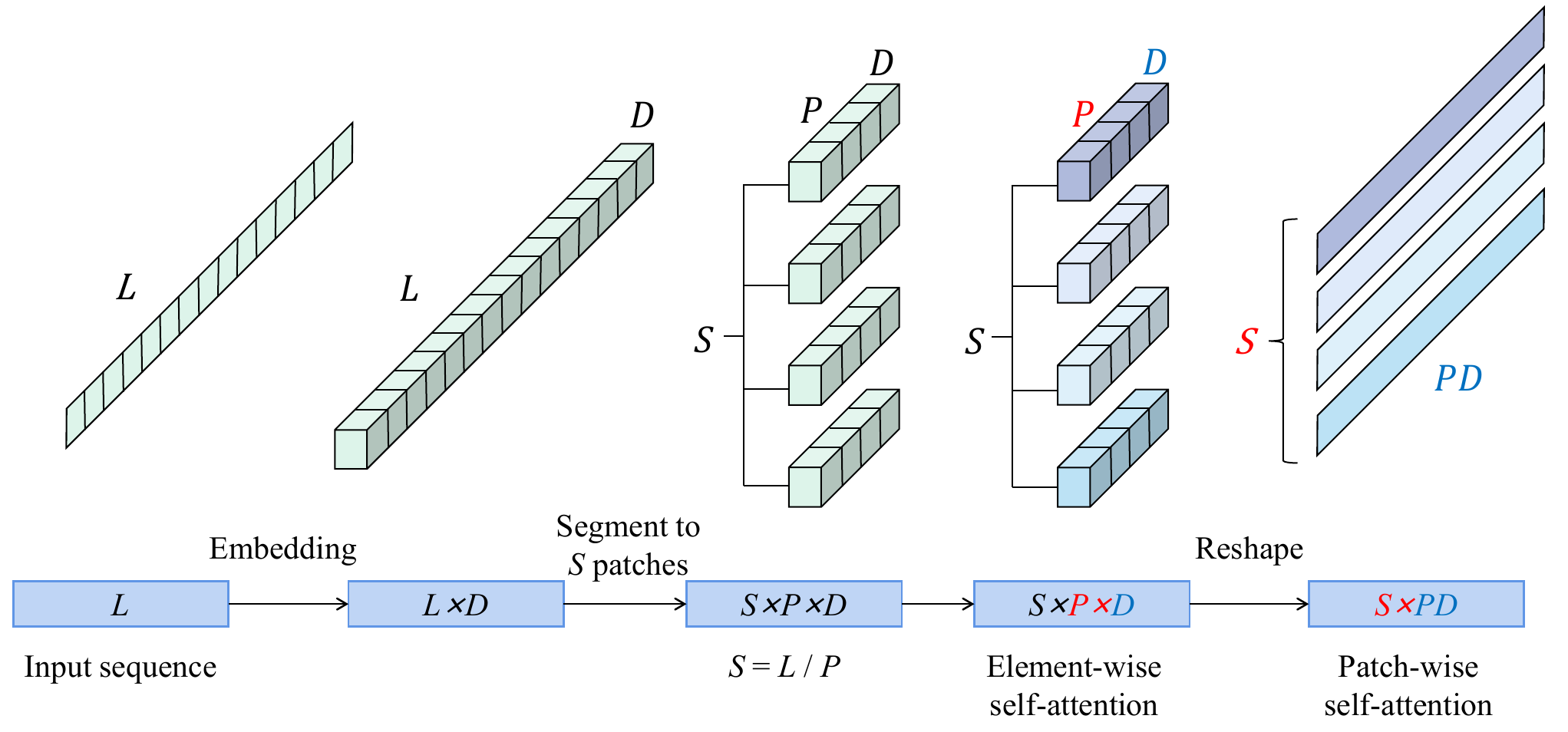}
\caption{The changes in the size of a single input sequence when propagating through the first encoder stage. The batch size and the variable dimension are omitted. The red and blue letters in the last two sizes separately refer to the token dimension and its latent representation dimension. The reshaping operation is used to treat the features of all elements in a single patch as a unity for the sake of connecting element-wise self-attention and patch-wise attention.}
\label{fig4}
\end{figure}

As shown in the schematic of Fig. \ref{fig2}, a element-wise attention block is arranged before each patch-wise block in every encoder stage to extract the inner-relationships of all patches before seeking their inter-relationship. This element-wise self-attention is patch-independent so that the additional complexity is $ O(P^2 \times L / P) = O(L \times P) $, which is linear with input sequence length. $ P $ is the patch size and $ L $ is the input sequence length in the last complexity expression. Observing that the element-wise attention requires the preservation of independent sequence elements information, therefore we cannot directly map the initially segmented patches into the latent space like other PTSFTs otherwise the element-wise information is no longer preserved and element-wise attention cannot be implemented. To address this issue, we adopt a channel-independent and element-independent embedding method. As illustrated in Fig. \ref{fig4} and Table \ref{tab1}, the input sequence elements of different timestamps do not interfere with each other during the embedding process and reshaping operations are performed for the different needs of tensor shapes of element-wise attention and patch-wise attention.\par
\begin{table}[!t]
\begin{center}
\caption{The architecture of the first stage in encoder}
\label{tab1}
\begin{tabular}{c|c}
\toprule[1.5pt]
Layer/Operation Name& Output Size \\
\midrule[1pt]
Input& $ L $\\
\hline
Embedding& $ L \times D $\\
\hline
Segmentation& $ S \times P  \times D, P = L / S $\\
\hline
Element-wise self-attention & $ S \times {\color[HTML]{FF0000} \textit{\textbf{P}}} \times { \textit{\textbf{D}}}  $  \\
\hline
Reshape& $ S \times PD  $ \\
\hline
Patch-wise self-attention& ${\color[HTML]{FF0000} \textit{\textbf{S}}} \times { \textit{\textbf{PD}}}$\\
\bottomrule[1.5pt]
\end{tabular}
\end{center}
\end{table}

The changes to decoder's attention are analogous but not completely. Since decoder itself has already owned two attention blocks, we maintain the patch-wise attention in cross-attention block to ensure the general construction of prediction sequence via auto-regressive process. Simultaneously, the masked self-attention block of vanilla TSFT is transformed to element-wise self-attention block, which is also patch-independent, in FPPformer. Just as we mentioned in the preceding sections, the prediction sequence is unknown so that we need to foremost guarantee the correctness of its general characteristics, manifested by placing patch-wise attention before the element-wise attention in decoder, then we can pursue the fine-grained features of prediction sequences without over-fitting. As the patch-wise cross-attention treats each patch as a unity, the respect to causality within prediction sequence, i.e., the masking to the upper triangular parts of query-key match matrix, is superfluous in the decoder of FPPformer.\par
\subsection{Diagonal-Masked (DM) Self-Attention}
\begin{figure}[!t]
	\centering
	\includegraphics[width=3.5in]{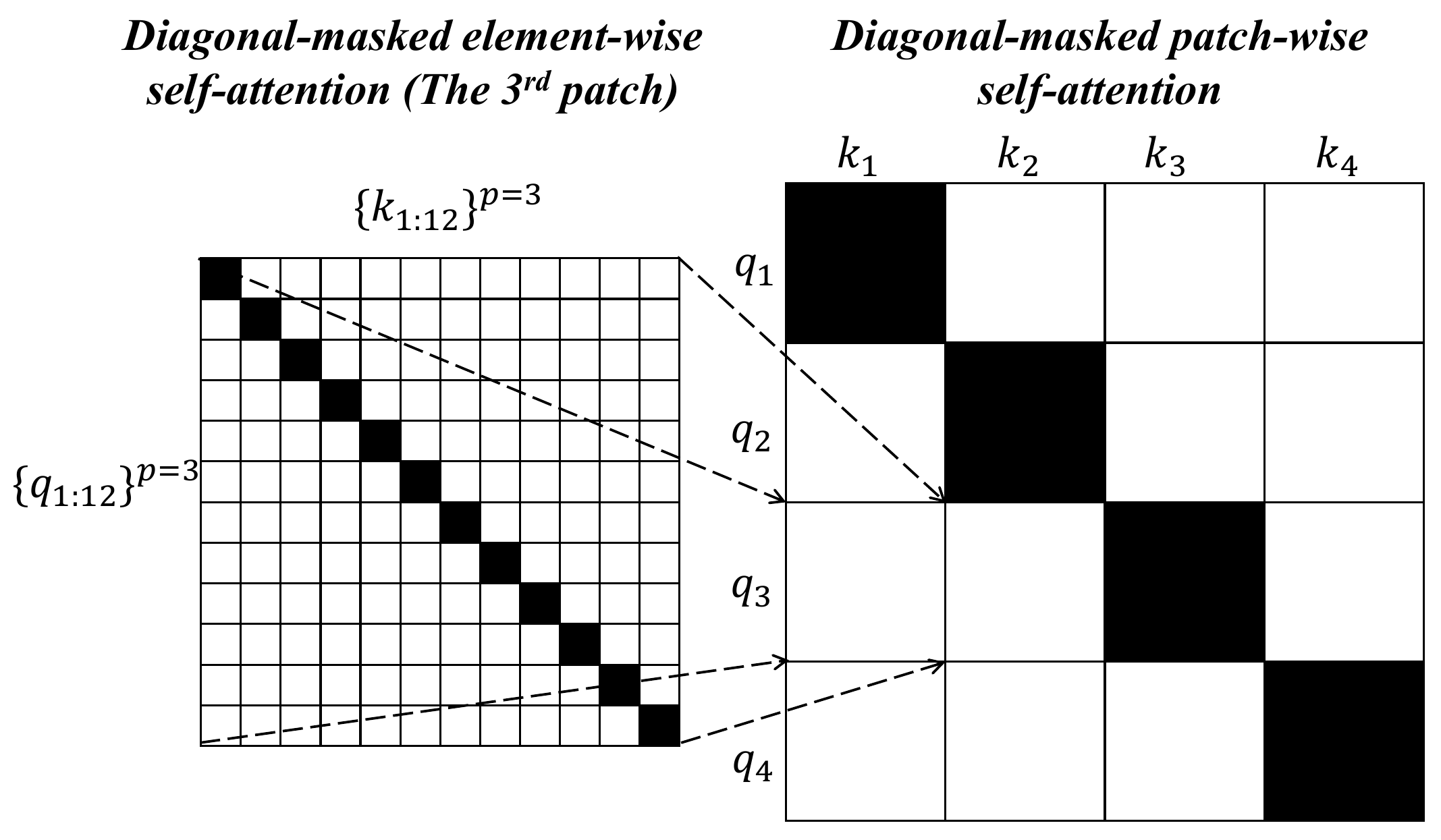}
	\caption{An example of query-key matching matrices in diagonal-masked element-wise self-attention and patch-wise self-attention. The total sequence length is 48 and the sequence is divided into 4 patches, each with the length of 12, in this example. The white hollow boxes denote the normal unchanged matrix elements while the black solid boxes, i.e., the matrix elements at the diagonal, denote the masked matrix elements.}
	\label{fig5}
\end{figure}

It is known that outliers always occur in real-world systems, especially for IoT systems owning immense and diverse data. These anomalies sometimes exist in the form of small patches \cite{Outlier} so that patch-attention cannot be immune to them. Compared with smoothing with filters \cite{NHiTS,FiLM}, which is natural but not flexible enough, it is better to devise mechanisms inside the networks to circumvent the negative effects of outliers in the latent space. Representation learning is a fair answer \cite{PatchTST} but needs heavy parameter tuning and not very stable. Directly masking the input sequence \cite{mask} gives rise to another problem analogous to the preceding position embedding problem in decoder since the fixed parameters cannot sufficiently represent the dynamic sequence features. Enlightened by \cite{UniAD}, we mask the diagonal of query-key matching matrix of both element-wise and patch-wise self-attention blocks in encoder, as sketched in Fig. \ref{fig5}. Thereby, any element(patch) during the attention can merely be expressed by the values of the rest of the elements(patches). Those elements or patches whose characteristics confine with the general ones are scarcely affected but the outliers are impossible to be expressed by normal elements(patches), hence their values are restored to approach the general level and the negative effects of them are mitigated.\par
\subsection{Projection}
The prediction sequence is acquired through the summation of the linear projections of encoder output and decoder output. The first linear projection is supposed to represent the linear correlations of input and prediction sequence while the second linear projection, together with the entire decoder, is supposed to represent the non-linear correlations. The loss function (\ref{eq2}) is the summation of MSE function (\ref{eq3}) and MAE function (\ref{eq4}) according to \cite{TS2Vec, CoST}.\par
\begin{small}
\begin{align}
Loss = \text{MSE}&(x_{t_2:t_3}^{1:V}, y_{t_2:t_3}^{1:V}) + \text{MAE}(x_{t_2:t_3}^{1:V}, y_{t_2:t_3}^{1:V})\label{eq2}\\
&\text{MSE}(x, y) = \frac{1}{n}\sum\nolimits_{i=1}^n (x_i-y_i)^2  \label{eq3}\\
&\text{MAE}(x,y) = \frac{1}{n}\sum\nolimits_{i=1}^n |x_i-y_i| \label{eq4}
\end{align}
\end{small}
\section{Experiments}
We attempt to answer three questions via the experiments on FPPformer:
\begin{enumerate}
	\item Can FPPformer outperform temporarily state-of-the-art TSFTs and TSFMs on commonly-used benchmarks with the settings of both short input sequence length and long input sequence length (Section \ref{Main_result})?
	\item Are the unique mechanisms proposed to be applied in FPPformer literally effective or useful (Section \ref{Ablation_study}) and what’s about their parameter sensitivity (Section \ref{Parameter_sensitivity})?
	\item Why does FPPformer own better or worse performances than other baselines? Can we figure it out via visualization (Section \ref{Case_study})?
\end{enumerate}
\subsection{Baselines and Datasets}
\begin{table}[!b]
	\begin{center}
		\caption{The numerical details of eight multivariate datasets}
		\label{tab2}
		\footnotesize
		\setlength\tabcolsep{10pt}
		\begin{tabular}{cccc}
			\toprule[1.5pt]
			Datasets  & Sizes  & Dimensions & Frequencies \\
			\midrule[1pt]
			ETTh$ _{1} $& 17420 & 7 & 1h\\
			ETTh$ _{2} $& 17420 & 7 & 1h\\
			ETTm$ _{1} $& 69680 & 7 & 15min\\	
			ETTm$ _{2} $& 69680 & 7 & 15min\\	
			ECL& 26304 & 321& 1h\\
			Traffic  & 17544 & 862& 1h\\		
			Weather &52696 &21 &10min	\\
			Solar &52560 &137 &10min	\\
			\bottomrule[1.5pt]
		\end{tabular}
	\end{center}
\end{table}
\begin{table}[]
	\begin{center}
		\caption{The numerical details of four M4 sub-datasets}
		\label{tab3}
		\footnotesize
		\setlength\tabcolsep{3pt}
		\begin{tabular}{ccccc}
			\toprule[1.5pt]
			Sub-datasets & Prediction Lengths &Periodicities& Frequencies &Instances \\
			\midrule[1pt]
			M4-Monthly   &  18 & 12  &1month  & 48,000\\
			M4-Weekly&  13 & 1   &1week  & 359   \\
			M4-Daily &  14 & 1   &1day  & 4,227 \\
			M4-Hourly & 48 & 24  &1hour  & 414   \\  
			\bottomrule[1.5pt]
		\end{tabular}
	\end{center}
\end{table}
\begin{table*}[!b]
	\begin{center}
		\caption{Multivariate forecasting results with short input length}
		\label{tab4}
		\footnotesize
		\setlength\tabcolsep{2.5pt}
		\begin{tabular}{c|c|cccc|cccc|cccc|cccc}
			\toprule[1.5pt]
			\multirow{2}{*}{Methods} & Datasets & \multicolumn{4}{c|}{ETTh$ _1 $}& \multicolumn{4}{c|}{ETTh$ _2 $}& \multicolumn{4}{c|}{ETTm$ _1 $}& \multicolumn{4}{c}{ETTm$ _2 $}\\
			\cmidrule(lr){2-2}
			\cmidrule(lr){3-6}
			\cmidrule(lr){7-10}
			\cmidrule(lr){11-14}
			\cmidrule(lr){15-18}
			&Metrics/Prediction Lengths& 96 & 192& 336& 720& 96 & 192& 336& 720& 96 & 192& 336& 720& 96 & 192& 336& 720\\
			\midrule[1pt]
			\multirow{2}{*}{FPPformer}   & MSE& \textit{\textbf{0.373}} & \textit{\textbf{0.425}} & \underline{\textit{0.470}}& \textit{\textbf{0.479}} & \textit{\textbf{0.296}} & \textit{\textbf{0.372}} & \underline{\textit{0.418}}& \textit{\textbf{0.422}} & \textit{\textbf{0.313}} & \textit{\textbf{0.362}} & \textit{\textbf{0.393}} & \textit{\textbf{0.448}} & \underline{\textit{0.176}}& \textit{\textbf{0.243}} & \textit{\textbf{0.302}} & \textit{\textbf{0.398}} \\
			& MAE& \textit{\textbf{0.391}} & \textit{\textbf{0.421}} & \textit{\textbf{0.442}} & \textit{\textbf{0.463}} & \textit{\textbf{0.343}} & \textit{\textbf{0.392}} & \textit{\textbf{0.427}} & \textit{\textbf{0.435}} & \textit{\textbf{0.350}} & \textit{\textbf{0.377}} & \textit{\textbf{0.401}} & \textit{\textbf{0.437}} & \textit{\textbf{0.258}} & \textit{\textbf{0.301}} & \textit{\textbf{0.340}} & \textit{\textbf{0.396}} \\
			\midrule
			\multirow{2}{*}{Triformer}   & MSE& 0.419 & 0.484 & 0.513 & 0.711 & 0.742 & 1.028 & 1.049 & 1.223 & 0.362 & 0.419 & 0.466 & 0.531 & 0.240 & 0.387 & 0.545 & 1.928 \\
			& MAE& 0.446 & 0.486 & 0.489 & 0.638 & 0.585 & 0.708 & 0.745 & 0.836 & 0.402 & 0.443 & 0.484 & 0.513 & 0.326 & 0.449 & 0.532 & 0.924 \\
			\multirow{2}{*}{Crossformer} & MSE& 0.472 & 0.533 & 0.698 & 0.847 & 0.884 & 2.835 & 2.339 & 4.387 & 0.444 & 0.522 & 0.649 & 0.818 & 0.350 & 0.643 & 1.051 & 3.988 \\
			& MAE& 0.493 & 0.516 & 0.602 & 0.713 & 0.690 & 1.381 & 1.241 & 1.810 & 0.467 & 0.524 & 0.592 & 0.672 & 0.421 & 0.596 & 0.774 & 1.426 \\
			\multirow{2}{*}{Scaleformer} & MSE& 0.443 & \underline{\textit{0.430}}& \textit{\textbf{0.457}} & 0.498 & 0.328 & 0.434 & 0.489 & 0.474 & \underline{\textit{0.327}}& 0.394 & 0.416 & \underline{\textit{0.467}}& \textit{\textbf{0.175}} & 0.251 & 0.316 & 0.438 \\
			& MAE& 0.460 & \underline{\textit{0.451}}& 0.472 & \underline{\textit{0.503}}& 0.380 & 0.448 & 0.490 & 0.495 & 0.381 & 0.422 & 0.441 & 0.474 & 0.268 & 0.319 & 0.361 & 0.436 \\
			\multirow{2}{*}{PatchTST}& MSE& \underline{\textit{0.393}}& 0.438 & 0.482 & \underline{\textit{0.492}}& \underline{\textit{0.296}}& \underline{\textit{0.380}}& \textit{\textbf{0.417}} & \underline{\textit{0.425}}& 0.332 & \underline{\textit{0.377}}& \underline{\textit{0.407}}& 0.473 & 0.177 & \underline{\textit{0.243}}& \underline{\textit{0.305}}& \underline{\textit{0.401}}\\
			& MAE& 0.427 & 0.459 & 0.494 & 0.519 & 0.346 & \underline{\textit{0.395}}& \underline{\textit{0.428}}& \underline{\textit{0.443}}& \underline{\textit{0.366}}& \underline{\textit{0.385}}& \underline{\textit{0.405}}& 0.442 & 0.274 & 0.324 & 0.363 & 0.420 \\
			\midrule
			\multirow{2}{*}{FiLM} & MSE& 0.423 & 0.474 & 0.510 & 0.557 & 0.299 & 0.384 & 0.425 & 0.443 & 0.354 & 0.390 & 0.422 & 0.481 & 0.183 & 0.249 & 0.310 & 0.409 \\
			& MAE& \textit{\underline{0.423}} & 0.452 & \underline{\textit{0.468}}& 0.521 & \underline{\textit{0.345}}& 0.397 & 0.436 & 0.454 & 0.371 & 0.387 & 0.408 & 0.441 & \underline{\textit{0.266}}& 0.307 & \underline{\textit{0.344}}& 0.399 \\
			\multirow{2}{*}{TSMixer} & MSE& 0.441 & 0.491 & 0.531 & 0.536 & 0.320 & 0.406 & 0.438 & 0.455 & 0.339 & 0.380 & 0.410 & 0.472 & 0.184 & 0.248 & 0.308 & 0.406 \\
			& MAE& 0.441 & 0.468 & 0.488 & 0.509 & 0.365 & 0.415 & 0.446 & 0.463 & 0.370 & 0.388 & 0.407 & \underline{\textit{0.441}}& 0.267 & \underline{\textit{0.307}}& 0.345 & \underline{\textit{0.399}}\\
			\midrule
			\midrule
			\multirow{2}{*}{Methods} & Datasets & \multicolumn{4}{c|}{ECL}  & \multicolumn{4}{c|}{Traffic} & \multicolumn{4}{c|}{Weather} & \multicolumn{4}{c}{Solar}\\
			\cmidrule(lr){2-2}
			\cmidrule(lr){3-6}
			\cmidrule(lr){7-10}
			\cmidrule(lr){11-14}
			\cmidrule(lr){15-18}
			&Metrics/Prediction Lengths& 96 & 192& 336& 720& 96 & 192& 336& 720& 96 & 192& 336& 720& 96 & 192& 336& 720\\
			\midrule[1pt]
			\multirow{2}{*}{FPPformer}   & MSE& \textit{\textbf{0.179}} & \textit{\textbf{0.186}} & \textit{\textbf{0.203}} & \textit{\textbf{0.246}} & \textit{\textbf{0.481}} & \textit{\textbf{0.479}} & \textit{\textbf{0.486}} & \textit{\textbf{0.518}} & \underline{\textit{0.174}}& \underline{\textit{0.219}}& 0.275& \textbf{\textit{0.352}} & 0.234 & 0.263 & 0.270 & 0.273 \\
			& MAE& \textit{\textbf{0.254}} & \textit{\textbf{0.264}} & \textit{\textbf{0.281}} & \textit{\textbf{0.317}} & \textit{\textbf{0.295}} & \textit{\textbf{0.290}} & \textit{\textbf{0.291}} & \textit{\textbf{0.309}} & \textit{\textbf{0.210}} & \textit{\textbf{0.251}} & \textit{\textbf{0.291}} & \textit{\textbf{0.342}} & 0.261 & 0.276 & \underline{\textit{0.277}} & 0.281 \\
			\midrule
			\multirow{2}{*}{Triformer}   & MSE& \underline{\textit{0.179}}& \underline{\textit{0.191}}& \underline{\textit{0.209}}& \underline{\textit{0.253}}&---&---&---&---& 0.174 & 0.219 & \underline{\textit{0.272}} & 0.357 & \underline{\textit{0.185}}& \underline{\textit{0.203}}& \underline{\textit{0.241}} & \underline{\textit{0.220}}\\
			& MAE& 0.277 & 0.289 & 0.308 & 0.344 &---&---&---&---& 0.242 & 0.290 & 0.323 & 0.378 & \underline{\textit{0.247}}& \underline{\textit{0.253}}& 0.285 & \underline{\textit{0.269}} \\
			\multirow{2}{*}{Crossformer} & MSE& 0.234 & 0.285 & 0.342 & 0.543 & \underline{\textit{0.532}}& \underline{\textit{0.543}}& \underline{\textit{0.562}} & \underline{\textit{0.571}}& \textit{\textbf{0.156}} & \textit{\textbf{0.210}} & \textit{\textbf{0.260}} & \underline{\textit{0.355}} & \textit{\textbf{0.167}} & \textit{\textbf{0.203}} & \textit{\textbf{0.216}} & \textit{\textbf{0.216}} \\
			& MAE& 0.323 & 0.351 & 0.381 & 0.471 & \underline{\textit{0.295}}& \underline{\textit{0.296}}& \underline{\textit{0.303}}& \underline{\textit{0.321}}& \underline{\textit{0.227}}& 0.278 & 0.321 & 0.385 & \textit{\textbf{0.214}} & \textit{\textbf{0.239}} & \textit{\textbf{0.251}} & \textit{\textbf{0.249}} \\
			\multirow{2}{*}{Scaleformer} & MSE& 0.180 & 0.195 & 0.215 & 0.257 & 0.551 & 0.578 & 0.587 & 0.611 & 0.198 & 0.313 & 0.395 & 0.595 & 0.206 & 0.218 & 0.266 & 0.283 \\
			& MAE& 0.297 & 0.308 & 0.328 & 0.363 & 0.341 & 0.350 & 0.354 & 0.364 & 0.273 & 0.372 & 0.420 & 0.549 & 0.305 & 0.303 & 0.330 & 0.333 \\
			\multirow{2}{*}{PatchTST}& MSE& 0.196 & 0.199 & 0.214 & 0.256 & 0.589 & 0.579 & 0.589 & 0.631 & 0.188 & 0.231 & 0.285 & 0.359 & 0.240 & 0.274 & 0.295 & 0.295 \\
			& MAE& 0.279 & 0.283 & \underline{\textit{0.299}}& \underline{\textit{0.331}}& 0.379 & 0.368 & 0.371 & 0.391 & 0.228 & \underline{\textit{0.263}}& \underline{\textit{0.301}}& \underline{\textit{0.349}}& 0.285 & 0.308 & 0.317 & 0.317 \\
			\midrule
			\multirow{2}{*}{FiLM} & MSE& 0.198 & 0.199 & 0.217 & 0.279 & 0.648 & 0.600 & 0.610 & 0.694 & 0.197 & 0.240 & 0.290 & 0.361 & 0.314 & 0.355 & 0.395 & 0.397 \\
			& MAE& \underline{\textit{0.274}}& \underline{\textit{0.278}}& 0.300 & 0.357 & 0.385 & 0.362 & 0.367 & 0.427 & 0.239 & 0.272 & 0.307 & 0.352 & 0.334 & 0.356 & 0.371 & 0.364 \\
			\multirow{2}{*}{TSMixer} & MSE& 0.222 & 0.226 & 0.244 & 0.287 & 0.725 & 0.706 & 0.727 & 0.775 & 0.191 & 0.235 & 0.286 & 0.358 & 0.263 & 0.302 & 0.332 & 0.335 \\
			& MAE& 0.308 & 0.315 & 0.332 & 0.363 & 0.453 & 0.449 & 0.458 & 0.477 & 0.236 & 0.271 & 0.307 & 0.352 & 0.305 & 0.328 & 0.345 & 0.345 \\
			\bottomrule[1.5pt]   
		\end{tabular}
	\end{center}
\end{table*}

To unveil the empirical forecasting capability of FPPformer, we perform multivariate forecasting experiments on eight benchmarks involved in four types of IoT systems, including electricity consumption (ETTh$ _1 $, ETTh$ _2 $, ETTm$ _1 $, ETTm$ _2 $ \cite{Informer}, ECL \cite{ECL}), traffic flow (Traffic \cite{Traffic}), meteorological conditions (Weather \cite{Weather}) and solar power production (Solar \cite{Solar}). Their numerical details are presented in Table \ref{tab2}. Eight temporarily state-of-the-art forecasting baselines, including four TSFTs (Triformer \cite{Triformer}, Crossformer \cite{Crossformer}, Scaleformer \cite{scaleformer}, PatchTST \cite{PatchTST}) and two TSFMs (FiLM \cite{FiLM} and TSMixer \cite{TSMixer}), are employed to make comparison with FPPformer. It is worth mentioning that they are all superb forecasting methods proposed in the recent two years. Specially, besides Scaleformer, the other three of four TSFTs are PTSFTs so that FPPformer does not have an edge on pure attention mechanism design. Furthermore, we notice that these six forecasting baselines own different variable treatment strategies, e.g., some of them are channel-independent while some of them are not, which means that multivariate forecasting results cannot fully typify their forecasting capabilities. Therefore, we additionally perform univariate forecasting results on M4 dataset \cite{M4}, which is a competition dataset qualified for univariate forecasting, rather than delibrately choosing a variate within the above multivariate forecasting datasets to perform univariate forecasting experiments like many other researches \cite{Triformer, Crossformer, scaleformer, FiLM}. Its details are elaborated in Table \ref{tab3}.\par
\subsection{Implementation Details}
\label{Implementation_details}

We would like to present a persuasive and fair comparison of FPPformer and other baselines, therefore we set the hyper-parameters of FPPformer identical to the commonly-used ones. The input sequence lengths are different in different sub-experiments but are kept identical for all baselines. The number of stages are 3 in both encoder and decoder of FPPformer, and the size of the initial segmented patch is 6, which are in accordance with those of Crossformer \cite{Crossformer}. The embedding dimension $ D $ (in Fig. \ref{fig4}) is 32. As for hyper-parameters with respect to the training process, FPPformer is trained via an Adam optimizer with the learning rate of 1e-4, which decays by half per epoch with totally ten epochs and the patience of one. The batch size is 16 and the dropout rate is 0.1. These are all commonly employed settings. All experiments, which are conducted on a single NVIDIA GeForce RTX 3090 24GB GPU, are repeated for five times with casual seeds and the average results are presented. The source codes are implemented by Python 3.8.8 and Pytorch 1.11.0 in \url{https://github.com/OrigamiSL/FPPformer}. {Correspondingly, the other baselines used in this work also merely employ the fixed hyper-parameters and settings, which are chosen after referencing their default ones. As for those with multiple choices and versions, we choose the one that owns the best general performance. How we use the other baselines for experiment can all be found in our provided GitHub repository.} The best results in each table are highlighted with \textbf{\textit{bold and italic}} and the second best are highlighted with \textit{\underline{underline and italic}}, barring a special table in Section \ref{Ablation_study}.\par
\subsection{Quantitative Results}
\label{Main_result}
\begin{table}[!t]
\begin{center}
\caption{Average multivariate results with long input lengths}
\label{tab5}
\footnotesize
\setlength\tabcolsep{6pt}
\begin{tabular}{c|c|ccc}
\toprule[1.5pt]
Methods                      & Metrics/Input Lengths & 192                     & 384                     & 576                     \\
\midrule[1pt]
\multirow{2}{*}{FPPformer}   & MSE (Avg. of all)                   & \textit{\textbf{0.358}} & \textit{\textbf{0.347}} & \textit{\textbf{0.345}} \\
& MAE (Avg. of all)                   & \textit{\textbf{0.354}} & \textit{\textbf{0.352}} & \textit{\textbf{0.350}} \\
\midrule
\multirow{2}{*}{Triformer}   & MSE (Avg. of all)                   & 0.932                   & 0.966                   & 1.546                   \\
& MAE (Avg. of all)                   & 0.621                   & 0.639                   & 0.828                   \\
\multirow{2}{*}{Crossformer} & MSE (Avg. of all)                   & 1.319                   & 1.192                   & 1.290                   \\
& MAE (Avg. of all)                   & 0.729                   & 0.707                   & 0.747                   \\
\multirow{2}{*}{Scaleformer} & MSE (Avg. of all)                   & 0.605                   & 0.555                   & 0.598                   \\
& MAE (Avg. of all)                   & 0.539                   & 0.517                   & 0.520                   \\
\multirow{2}{*}{PatchTST}    & MSE (Avg. of all)                   & \underline{\textit{0.377}}    & \underline{\textit{0.357}}    & \underline{\textit{0.354}}    \\
& MAE (Avg. of all)                   & \underline{\textit{0.378}}    & \underline{\textit{0.368}}    & \underline{\textit{0.366}}    \\
\midrule
\multirow{2}{*}{FiLM}        & MSE (Avg. of all)                   & 0.397                   & 0.381                   & 0.383                   \\
& MAE (Avg. of all)                   & 0.391                   & 0.385                   & 0.390                   \\
\multirow{2}{*}{TSMixer}     & MSE (Avg. of all)                   & 0.397                   & 0.372                   & 0.367                   \\
& MAE (Avg. of all)                   & 0.392                   & 0.379                   & 0.377                  
\\
			\bottomrule[1.5pt]  
		\end{tabular}
	\end{center}
\end{table}

We commence with the multivariate forecasting experiments, whose results are shown in Table \ref{tab4} and Table \ref{tab5}. Under many real-world occasions, training samples are limited so that long input sequence length is not always available for some deep forecasting methods needing it for satisfactory performances. Therefore, we first measure the performances of FPPformer and its six competitors in eight multivariate benchmarks with input sequence length of 96, which is ascribable to well-known Autoformer \cite{Autoformer} in Table \ref{tab4}. The prediction lengths are commonly agreed-upon \{96, 192, 336, 720\}. Then we evaluate the performances of the same seven models and datasets using longer input sequence length within \{192, 384, 576\}, whose results are shown in Table \ref{tab5}. MSE (\ref{eq3}) and MAE (\ref{eq4}) are utilized as the evaluation metrics. The average results of eight benchmarks with prediction length of 720 are given in Table \ref{tab5} to refrain from tedious data stacking. Full results are given in our released repository provided in Section \ref{Implementation_details}. `---' refers to the fact that certain model is out of the memory (24GB) even batch size is set to 1.\par
As Table \ref{tab4} and \ref{tab5} show, FPPformer outperforms other baselines in most of situations with both short and long input sequence lengths. When input sequence length is set to 96, FPPformer obtains 31.7\%/60.0\%/10.5\%/6.4\%/12.8\%/14.7\% MSE reduction compared with Triformer/Crossformer/ Scaleformer/PatchTST/FiLM/TSMixer, which illustrates the superb forecasting capability of FPPformer with the setting of short input sequence length. Though it seems that FPPformer fails to own a superior performance when experimenting on Solar dataset, FPPformer reconquers its leading position with longer input length when handling the same dataset (Concrete results are available at github repository provided in Section \ref{Implementation_details}). {Furthermore, if also equiped with cross-variable attention\footnote{We use the statement of `cross-variable', rather than `cross-dimension' in Crossformer, to maintain the identical description of the variable dimension throughout this work.} like Crossformer, which means that a cross-variable attention module proposed by Crossformer is arranged at the end of each stage of the encoder and decoder in FPPformer, the modified FPPformer, denoted by FPPformer-Cross in Table \ref{tab_cross}, is capable of completely outperforming Crossformer and iTransformer \cite{iTransformer}, which is another state-of-the-art model employing cross-variable attention, under Solar dataset.}\par 
\begin{table}[]
	\begin{center}
		\caption{Comparison of models employing cross-variable attention}
		\label{tab_cross}
		\footnotesize		
		\setlength\tabcolsep{0.8pt}
		\begin{tabular}{c|c|cccc}
			\toprule[1.5pt]
			Methods (Dataset: Solar)                          & Metrics/Prediction Length & 96                      & 192                     & 336                     & 720                     \\
			\midrule
			\multirow{2}{*}{FPPformer-Cross} & MSE                       & \textit{\textbf{0.162}} & \textit{\textbf{0.174}} & \textit{\textbf{0.191}} & \textit{\textbf{0.199}} \\
			& MAE                       & \textit{\textbf{0.209}} & \textit{\textbf{0.232}} & \textit{\textbf{0.237}} & \textit{\textbf{0.238}} \\
			\midrule
			\multirow{2}{*}{iTransformer}    & MSE                       & 0.215                   & 0.247                   & 0.267                   & 0.268                   \\
			& MAE                       & 0.255                   & 0.279                   & 0.294                   & 0.297                   \\
			\multirow{2}{*}{Crossformer}     & MSE                       & \underline{\textit{0.167}}    & \underline{\textit{0.203}}    & \underline{\textit{0.216}}    & \underline{\textit{0.216}}    \\
			& MAE                       & \underline{\textit{0.214}}    & \underline{\textit{0.239}}    & \underline{\textit{0.251}}    & \underline{\textit{0.249}}   
			 
			\\   
			\bottomrule[1.5pt]                
		\end{tabular}
		
	\end{center}
\end{table}

When prolonging the input sequence length within \{192, 384, 576\}, FPPformer achieves better forecasting performances and respectively obtains 9.5\%/12.3\%/13.0\% general MSE reduction when compared with the FPPformer with short input length in Table \ref{tab4}. Moreover, FPPformer obtains 67.4\%/72.3\%/40.2\%/3.5\%/9.5\%/7.5\% MSE reduction compared with Triformer/Crossformer/Scaleformer/PatchTST/ FiLM/TSMixer with longer input lengths in general. These phenomena illustrate the superb forecasting performances of FPPformer especially handling longer input sequence.\par
\begin{table}[!b]
	\begin{center}
		\caption{Univariate forecasting results}
		\label{tab6}
		\footnotesize
		\setlength\tabcolsep{2pt}
		\begin{tabular}{c|c|cccc}
			\toprule[1.5pt]
			Methods                      & Metrics & M4-Monthly              & M4-Weekly               & M4-Daily                & M4-Hourly                \\
			\midrule[1pt]
			\multirow{2}{*}{FPPformer}   & SMAPE   & \textit{\textbf{8.674}} & \textit{\textbf{8.711}} & \textit{\textbf{3.214}} & \textit{\textbf{17.468}} \\
			& OWA     & \textbf{\textit{0.891}}    & \underline{\textit{1.464}}    & \textit{\textbf{1.057}} & \textit{\textbf{1.373}}  \\
			\midrule
			\multirow{2}{*}{Triformer}   & SMAPE   & 65.161                  & 106.905                 & 82.288                  & 78.169                   \\
			& OWA     & 11.068                  & 19.181                  & 41.565                  & 5.756                    \\
			\multirow{2}{*}{Crossformer} & SMAPE   &   24.801  & 12.702                  & 6.901                   & 28.471                   \\
			& OWA     &  2.704 & 3.021                   & 3.077                   & 2.706                    \\
			\multirow{2}{*}{Scaleformer} & SMAPE   & 14.275 & \underline{\textit{10.591}}   & \underline{\textit{3.362}}    & 19.939    \\
			& OWA     & 1.163& \textit{\textbf{1.409}} & 1.163                   & 2.017                    \\
			\multirow{2}{*}{PatchTST}    & SMAPE   & 15.127                  & 15.687                  & 4.457                   & 47.657                   \\
			& OWA     & 1.237                   & 2.679                   & 1.519                   & 7.526                    \\
			\midrule
			\multirow{2}{*}{FiLM}        & SMAPE   & 32.329                  & 15.029                  & 8.376                   & 21.681                   \\
			& OWA     & 3.489                   & 3.966                   & 3.899                   & 2.240                    \\
			\multirow{2}{*}{TSMixer}     & SMAPE   & \underline{\textit{13.886}}   & 10.962                  & 3.429                   & \underline{\textit{19.769}}                   \\
			& OWA     & \underline{\textit{0.947}}                   & 1.515                   & \underline{\textit{1.131}}    & \underline{\textit{1.516}}   \\
			\bottomrule[1.5pt]            
		\end{tabular}	
	\end{center}
\end{table}

\begin{table*}[]
	\begin{center}
		\caption{Ablation results with prediction lengths of 720}
		\label{tab7}
		\footnotesize		
		\setlength\tabcolsep{0.8pt}
		\begin{tabular}{c|c|c|c|cccccccc|c}
			\toprule[1.5pt]
			Architecture               & Attention                 & Decoder Sturcture            & DM & ETTh$ _1 $ & ETTh$ _2 $ & ETTm$ _1 $ & ETTm$ _2 $ & ECL   & Traffic & Weather & Solar & Avg. MSE(Increase rate) \\
			\midrule[1pt]
			\multirow{8}{*}{FPPformer} & Point-wise                & top-down                     &\checkmark & 0.706 & 0.446 & 0.708 & 0.404 & 0.818 & 1.298   & 0.343   & 0.371 & 0.637(+84.5\%)    \\
			& Patch-wise                & top-down                     &\checkmark  & 0.479 & 0.422 & 0.441 & 0.384 & 0.227 & 0.501   & 0.336   & 0.248 & 0.380(+10.0\%)    \\
			& Point-wise \& Patch-wise & \thead{bottom-up\\(e.g. Crossformer)} &\checkmark& 0.471 & 0.427 & 0.446 & 0.405 & 0.235 & 0.522   & 0.340   & 0.264 & 0.389(+12.6\%)    \\
			& Point-wise \& Patch-wise & \thead{Linear\\(e.g. PatchTST)}       &\checkmark & 0.473 & 0.425 & 0.437 & 0.385 & 0.229 & 0.488   & 0.238   & 0.333 & 0.376(+9.0\%)    \\
			& Point-wise \& Patch-wise & top-down                     &$ \times $ & 0.469 & 0.421 & 0.439 & 0.385 & 0.227 & 0.500   & 0.337   & 0.254 & 0.379(+9.8\%)    \\
			(Complete FPPformer)       & Point-wise \& Patch-wise & top-down                     &\checkmark & 0.449 & 0.394 & 0.411 & 0.366 & 0.199 & 0.432   & 0.316   & 0.194 & 0.345   
			\\   
			\bottomrule[1.5pt]                
		\end{tabular}
		
	\end{center}
\end{table*}

Then we compare the univariate forecasting capability of FPPformer with other six baselines on M4. We omit the first two subsets with sampling frequencies of a year and a quarter since many of their instance lengths are too short, and only perform experiments on the rest of four subsets. The prediction lengths, which are regulated by \cite{M4}, are \{18, 13, 14, 48\} for \{M4-Monthly, M4-Weekly, M4-Daily, M4-Hourly\}. The input sequence lengths for them are separately \{72, 65, 84, 336\} after consulting \cite{NBEATS}. We change these four input sequence lengths a little to \{72, 72, 96, 384\} for the sake of rendering them fitting the patch-wise attention in FPPformer. The M4-specifical metrics SMAPE (\ref{eq5}) and OWA (\ref{eq7}) are used for measurement. $ m $ refers to the periodicity of series and naïve2 refers to the results of a seasonally-adjusted forecast model by \cite{M4} for scaling in OWA.\par
\begin{small}
	\begin{equation}
		\label{eq5}
		\text{SMAPE} = \frac{200}{t_3 - t_2}\sum_{t\in [t_2,t_3)}\frac{|y_t-x_t|}{|y_t| + |x_t|}
	\end{equation}
\end{small}
\begin{small}
	\begin{equation}
		\label{eq6}
		\text{MASE} = \frac{1}{t_3\!-\! t_2}\sum_{t\in [t_2,t_3)}\frac{|y_t \! -\! x_t|}{\frac{1}{t_3 \!-\! t_1 \!-\! m}\sum_{j\in [t_1\!+\! m\! +\! 1,t_3)}|x_j \! -\! x_{j-m}|}
	\end{equation}
\end{small}
\begin{small}
	\begin{equation}
		\label{eq7}
		\text{OWA} = \frac{1}{2}(\frac{\text{SMAPE}}{{\text{SMAPE}_{\text{Naïve2}}}} + \frac{\text{MASE}}{{\text{MASE}_{\text{Naïve2}}}})
	\end{equation}
\end{small}

The univariate forecasting results are shown in Table \ref{tab6}. It could be observed that FPPformer gains 88.6\%/47.8\%/ 21.0\%/54.1\%/50.8\%/20.8\% MSE reduction in general when compared with Triformer/Crossformer/Scaleformer/PatchTST/ FiLM/TSMixer, which is more persuasive to verify the forecasting capability of FPPformer.\par
\begin{table}[!t]
	\begin{center}
		\caption{MSE results of parameter sensitivity on stage numbers}
		\label{tab8}
		\footnotesize		
		\setlength\tabcolsep{2.5pt}
		\begin{tabular}{c|cccc|c}
			\toprule[1.5pt]
			Methods/Stages & $ M_1(\tilde{M_1}) $                      & $ M_2(\tilde{M_2}) $                       & $ M_4(\tilde{M_3}) $                       & $ M_4(\tilde{M_4}) $        & Dev \\
			\midrule[1pt]
			FPPformer  & \textit{\textbf{0.349(1.00)}} & \textit{\textbf{0.344(0.99)}} & \textit{\textbf{0.345(0.99)}} & \textit{\textbf{0.351(1.01)}} &  \textit{\textbf{0.03}} \\
			Scaleformer & 0.417(1.00) & 0.425(1.02) & 0.434(1.04) & 0.433(1.04) & 0.10                   \\
			Triformer   & 0.787(1.00)            & 0.797(1.01)                   & 0.964(1.23)                   & 0.814(1.03)                  & 0.27                   \\
			PatchTST    &\underline{\textit{0.370(1.00)}} & \underline{\textit{0.369(1.00)}} & \underline{\textit{0.385(1.04)}}    &  \underline{\textit{0.382(1.03)}}    & \underline{\textit{0.07}}    \\
			Crossformer & 1.475(1.00)               &  1.339(0.91)                   & 1.466(0.99)                   & 1.498(1.02)                   &  0.12                 \\
			\bottomrule[1.5pt]    
		\end{tabular}
		
	\end{center}
\end{table}
\begin{figure*}[!b]
	\centering
	\includegraphics[width=6in]{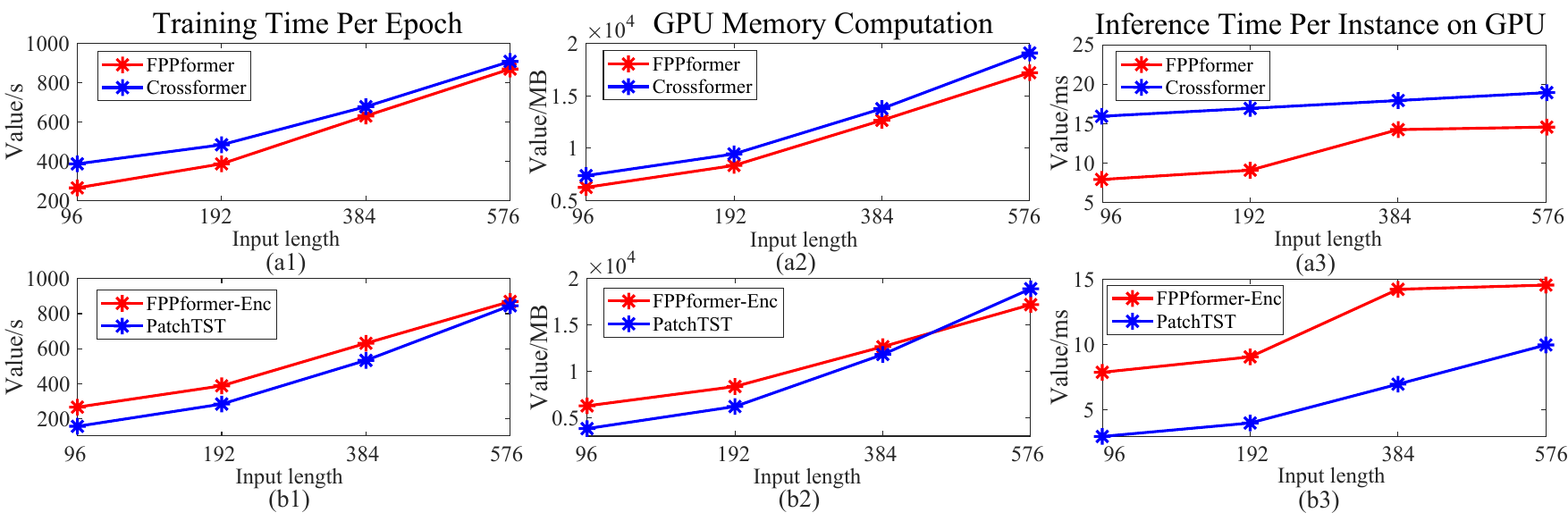}
	\caption{{ The results of training time per epoch (1), GPU memory computation (2) and inference time per instance (3) in two experiments (a)(b).}}
	\label{fig11}
\end{figure*}

\subsection{Ablation Study}
\label{Ablation_study}
We conduct ablation studies to validate the functions of the architecture of FPPformer and its components. All variants are experimented with multivariate benchmarks with prediction sequence length of 720. The results of all eight benchmarks are presented in Table \ref{tab7}. Each value is the average of four sub-experiment results with input sequence lengths of \{96, 192, 384, 576\}. As expected, only using point-wise attention like canonical TSFT gives rise to a losses increasing of 84.6\% (Average loss: \textbf{0.345}$ \rightarrow $0.637). Meanwhile, only using patch-wise attention like other PTSFTs does not suffer a severe performance degradation but still owns apparently worse performance than our proposed combined patch-wise attention and point-wise attention (Average loss: 0.380 vs. \textbf{0.345}). As for the decoder architecture design, FPPformer surpasses the same models but replacing the decoder architecture of FPPformer with that of Crossformer (Average loss: \textbf{0.345} vs. 0.389) or simply substituting the entire decoder with a linear projection like PatchTST (Average loss: \textbf{0.345} vs. 0.376). Besides, removing DM mechanisms in the self-attention blocks of encoder also results in worse performances (Average loss: \textbf{0.345}$ \rightarrow $0.379). Conclusively, the efficiency and necessity of all unique parts and architectures of FPPformer are verified.\par
\subsection{Parameter Sensitivity Analysis}
\label{Parameter_sensitivity}
It is well-known that TSFT cannot own too many layers or stages, otherwise the risk of over-fitting substantially rises. Thereby, we perform parameter analysis on the stage number of FPPformer in this section to check out whether FPPformer is capable of handling this problem. The parameter analysis of patch size is no longer tested as it has been well studied by \cite{Triformer,Crossformer,PatchTST}. The input sequence length is chosen as 576 for using more stages for FPPformer and other models use their default input sequence lengths, which are supposed to be best for them (96 for \{Scaleformer, Triformer\}; 512 for PatchTST; 336 for Crossformer). The number of stages are chosen in \{1, 2, 3, 4\}. The prediction sequence length is set as 720 and the average results (MSEs) of all eight multivariate benchmarks are presented in Table \ref{tab8}. The result of each model with stage number of one is utilized as the normalization factor to further measure their absolute performance deviations with more stage numbers which are manifested by (\ref{eq8}).\par
\begin{small}
	\begin{equation}
		\label{eq8}
		\text{Dev} = \sum\nolimits_{i=2}^4 |\tilde{M_i} - \tilde{M_1}|, \tilde{M_i} = M_i / M_1
	\end{equation}
\end{small}

$ M_i (i=1,2,3,4) $ are the original average MSE results and $ \tilde{M_i} (i=1,2,3,4) $ are the normalized ones. When comparing the performances and performance deviations of five TSFTs, it is evident that FPPformer not only keeps its leading position (smaller errors) among all TSFTs with different stage numbers, but also maintains its robustness ascendancy (smaller deviations) over other TSFTs.\par
\subsection{{Complexity Analysis}}
\begin{figure*}[]
	\centering
	\includegraphics[width=5.5in]{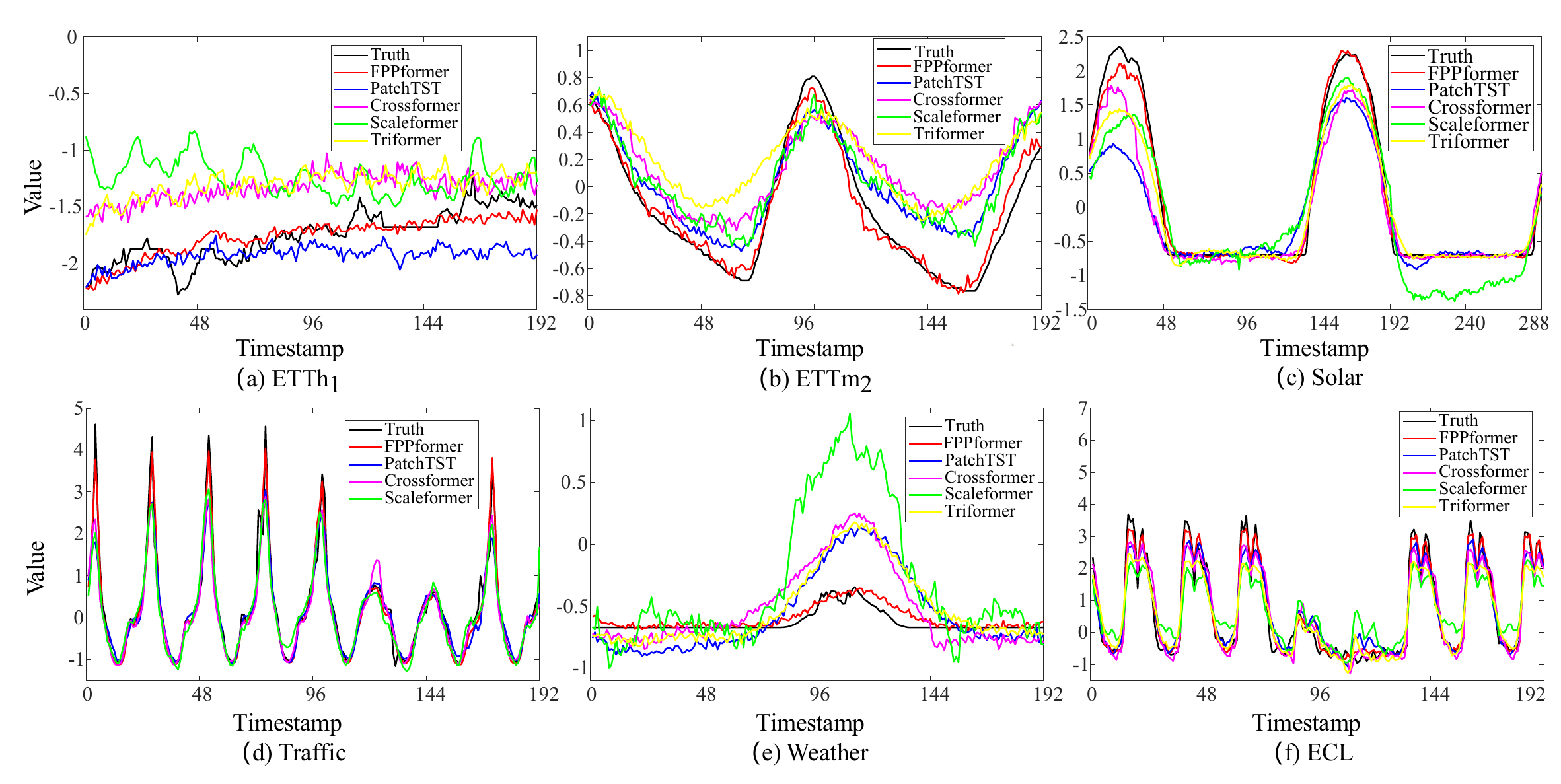}
	\caption{Forecasting windows of FPPformer and other four TSFTs from six datasets. The black line is the ground truth, the red line is the forecasting curve of FPPformer and the rest are the forecasting curves of other TSFTs.}
	\label{fig6}
\end{figure*}

{{We compare the training time per epoch/the GPU memory computation/the inference time per instance on GPU of FPPformer and Crossformer \cite{Crossformer}, and the identical four measuring criteria of FPPformer without decoder and PatchTST \cite{PatchTST} during multivariate forecasting under Solar dataset. Solar dataset is selected in that it owns the intermediate variable number among all datasets. Crossformer and PatchTST are chosen as they are also patch-wise attention based models. The decoder of FPPformer is removed when compared with PatchTST since PatchTST does not employ the decoder architecture. The input sequence lengths are chosen within \{96, 192, 384, 576\} and the prediction sequence length is 96. The other hyper-parameters and settings are identical with those used in the quantitative multivariate results, barring the size of the hidden(embedding) dimension. The size of the hidden dimension, which is the one that exceedingly affects the model complexity, is identical for all baselines in this experiment so that the model architecture design can determine the model complexity to the utmost extent. These two experiment results are shown in Fig. \ref{fig11}(a)(b) respectively.\par
As Fig. \ref{fig11}(a) shows, the full FPPformer owns linear computation and space complexity with input sequence length. Moreover, Fig. \ref{fig11}(b) illustrates that the encoder of FPPformer (FPPformer-Enc for short) also only owns linear complexity and the complexity of FPPformer and PatchTST are analogous, demonstrating that the element-wise attention, which is almost the only difference between PatchTST and FPPformer-Enc, merely brings minuscule additional complexity.} \par
\subsection{Case Study}
\label{Case_study}
\begin{figure}[!b]
	\centering
	\includegraphics[width=3.5in]{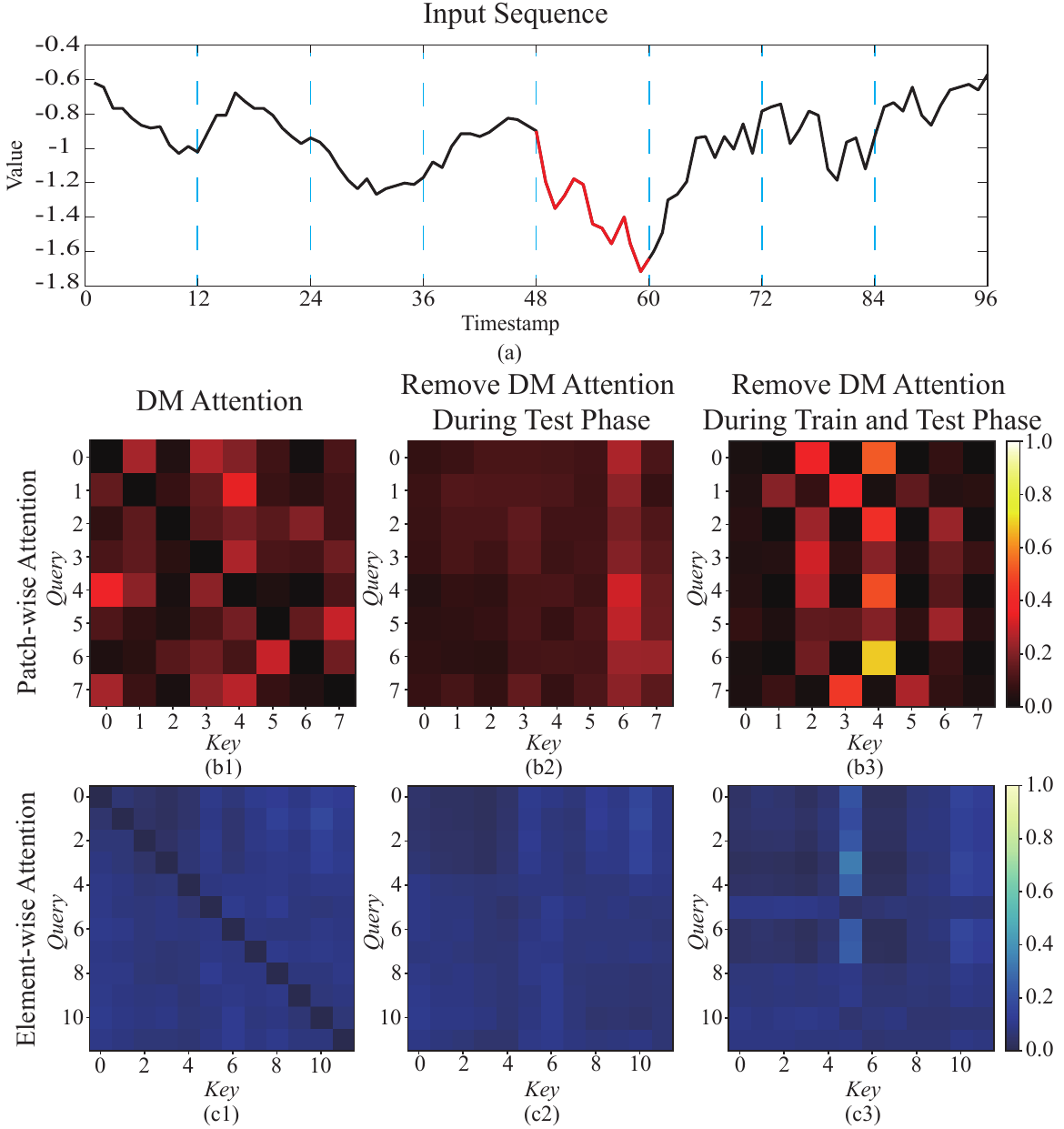}
	\caption{{ The visualization of DM patch-wise (b) and element-wise (c) self-attention score distributions via heat map. All figures are obtained by FPPformer or its variants with the identical input sequence (a). Specially, the element-wise attention score heat maps stem from the fifth patch, which is supposed to be anomalous and marked with red in (a).}}
	\label{fig7}
\end{figure}

To more vividly illustrate the outstanding forecasting performance of FPPformer, we visualize several forecasting windows of FPPformer and other TSFTs from different datasets in Fig. \ref{fig6}. Benefitting from exploiting better decoder and attention mechanism, FPPformer excels in capturing the features of trends (Fig. \ref{fig6}(a)), seasons (Fig. \ref{fig6}(b)) and their hybrids (Fig. \ref{fig6}(c)) so that smaller forecasting errors than others can be obtained. Moreover, its preponderance of robustness and immunity against outliers are revealed in Fig. \ref{fig6}(d)(e)(f) where certain distribution shifts occur in partial distincts. {In addition, we present some visualizations of the feature maps of FPPformer and several competitors in the latent space to validate the functions of its unique modules.}\par
{a. We visualize the attention score distribution of the first DM patch-wise and DM element-wise attention in FPPformer with a certain input window of length 96 in ETTh$ _1 $ dataset, via heat map in Fig.\ref{fig7}. As illustrated in Fig. \ref{fig7}(b1), the attention} {score distribution is uniformly distributed if applying the DM patch-wise attention during the training phase. Even substituting the DM patch-wise attention with the normal attention solely in the testing phase (Fig.\ref{fig7}(b2)) will not lead to the self-matches with high scores, demonstrating the enhancement of DM attention mechanism on universal feature extraction. However, it can be observed in Fig.\ref{fig7}(b3) that the highest attention score chiefly lies in the fifth patch, which corresponds to an outlier patch with a exorbitant dip. The visualization (Fig.\ref{fig7}(c)) of the three different element-wise attention score heat maps of that outlier patch in Fig.\ref{fig7}(b) also manifests the same phenomenon,i.e., the excessive high or low values can result in the excessive high attention scores which burden the universal feature extraction.}\par
\begin{figure}[!b]
	\centering
	\includegraphics[width=3.5in]{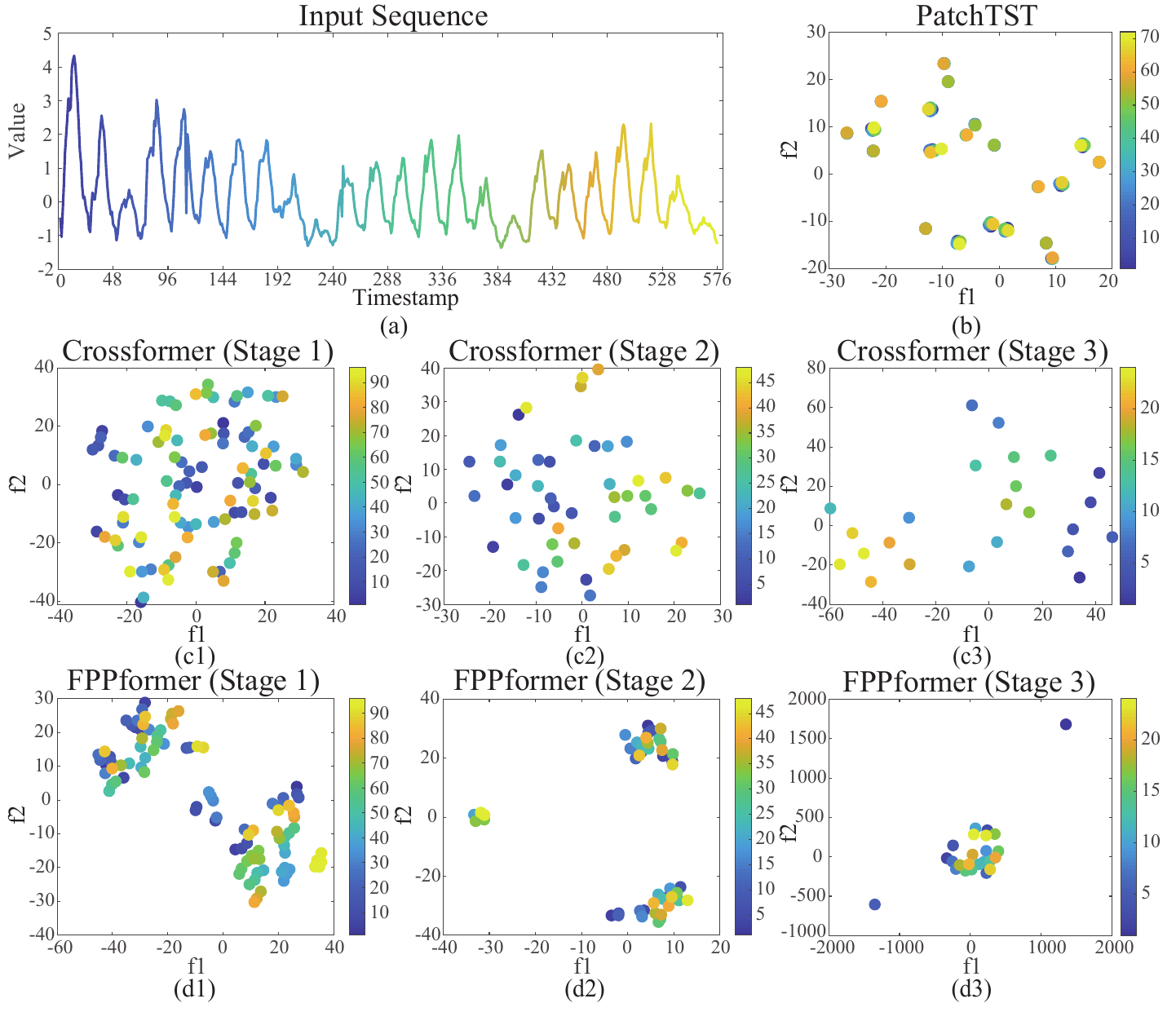}
	\caption{{ The visualization of the feature maps of the patches in different encoder stages (a) in PatchTST (b), Crossformer (c) and FPPformer (d) via T-SNE. The points with different colors denote different patches. Only the feature map of the last encoder stage of PatchTST is shown since it is not hierarchical.}}
	\label{fig8}
\end{figure}

{b. As the additional element-wise attention is employed in FPPformer to reinforce the inner-patch feature extraction, we visualize the feature maps outputted by each patch-wise attention in the encoders of FPPformer, PatchTST and Crossformer by T-SNE \cite{tsne} in Fig.\ref{fig8}. To get rid of the influences from other different modules, this experiment is performed as a reconstruction experiment, which reconstructs the input sequences via the ultimate encoder features, and we remove the DM technique in the patch-wise attentions of FPPformer but keep the DM element-wise attention. The input sequence length is set as 576 to provide enough data points in the T-SNE figures. Apparently, the distances of the data points in three FPPformer-related figures are the smallest among three models, indicating that  the DM element-wise attentions can literally assist in better representing the features of each patch in the latent space so that it is easier for the following patch-wise attentions in FPPformer encoder to extract the universal features among different patches and FPPformer can have an edge on extracting universal features compared with other two baselines.}\par
\begin{figure}[]
	\centering
	\includegraphics[width=3.5in]{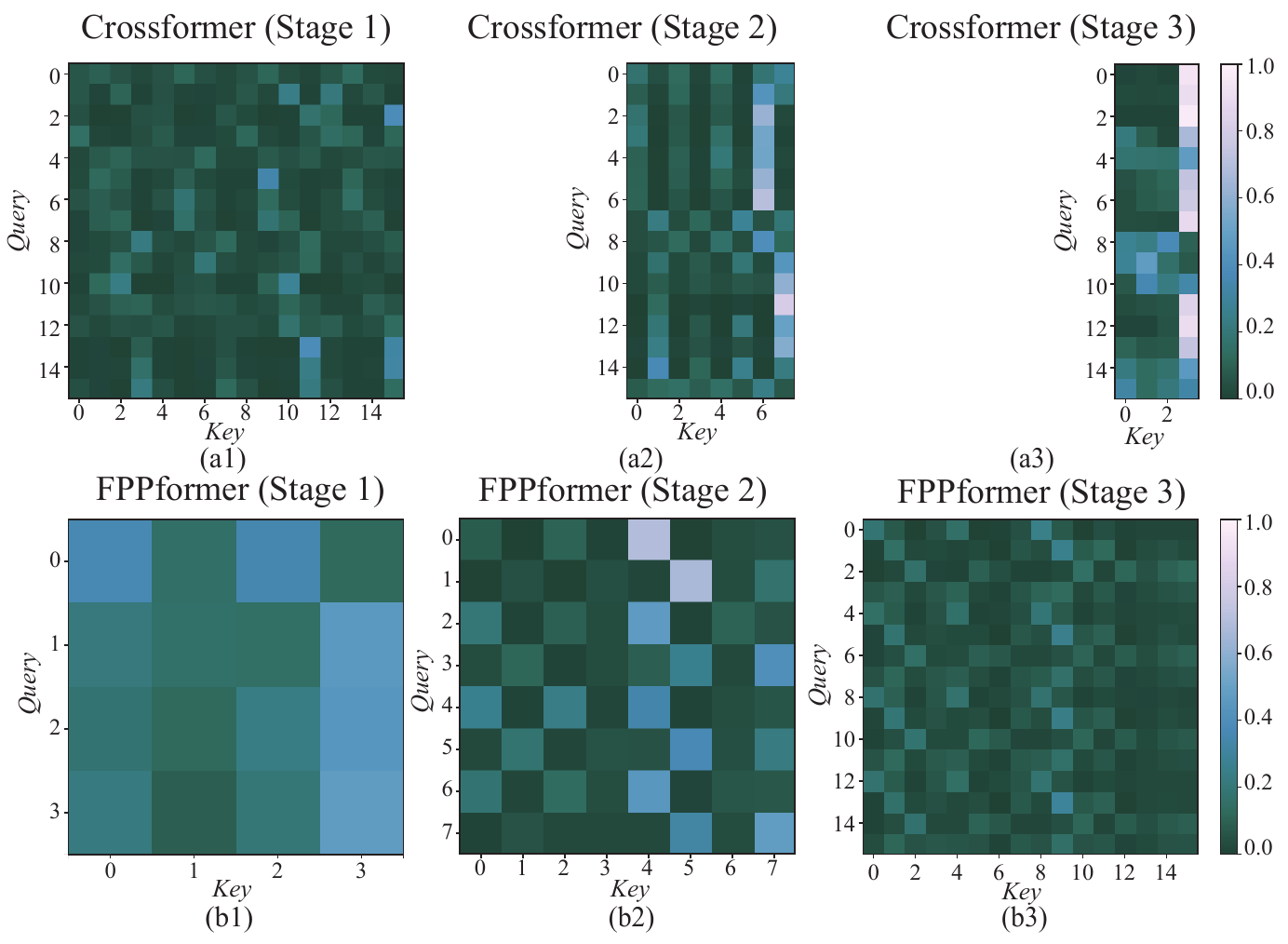}
	\caption{{ The visualization of the cross-attention score distribution of different decoder stages in Crossformer (a) and FPPformer (b) via heat map.}}
	\label{fig9}
\end{figure}
\begin{figure}[!b]
	\centering
	\includegraphics[width=3.5in]{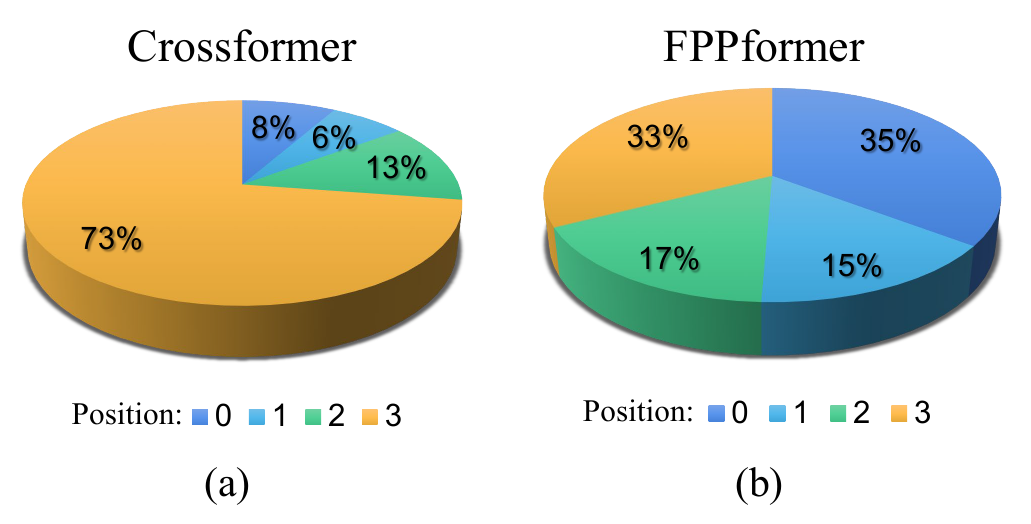}
	\caption{{ The distribution of the highest cross-attention score positions of the third stage in Crossformer (a) and the first stage in FPPformer (b).}}
	\label{fig10}
\end{figure}

{c. To vividly illustrate that the top-down architecture in decoder can genuinely render the construction of prediction sequence feature maps more general in the latent space, we visualize the patch-wise cross-attention score distributions of different decoder layers in FPPformer and Crossformer via heat maps in Fig. \ref{fig9}. Obviously, with the stage number grows, the attention score matrix size is getting bigger in FPPformer and smaller in Crossformer, illustrating the differences between `bottom-up' and `top-down' decoder architecture. Meanwhile, it can be observed that the highest attention scores in Crossformer primarily locate at the last patch, especially the third stage, indicating that the construction of the prediction sequence features heavily rely on the rear end of input sequence features and it fails to build up the prediction sequence in an universal manner. On contrary, the cross-attention scores in the decoder uniformly distribute along the temporal dimension, which implies the preponderance of the top-down architecture in FPPformer decoder. In effect, the instance in Fig. \ref{fig9}. is not a particular situation. We collect the highest attention score positions of the third stage in Crossformer and the first stage in FPPformer, where the most coarse-grained features lie in, when handling the whole ETTh$ _1 $ dataset, i.e., with over 100, 000 instances. The result is shown in Fig.\ref{fig10}. Apparently, the highest attention score distribution of FPPformer is much more uniform than that of Crossformer, indicating the success of the top-down decoder design in FPPformer.}\par
\section{{Discussion}}
{Though FPPformer has achieved state-of-the-art performances, it still owns at least two limits:\par
1. The hierarchy in FPPformer can be more exquisitely devised. The `merging' operation in the encoder of FPPformer is too simple to well represent the feature map of the bigger patch via the two smaller patch ingredients. So does the `splitting' operation in the decoder. The cutting-edge methods to handle the combination or the split of patches, e.g., SwinTransformer \cite{SwinTransformer}, in CV field, where patch-wise attention is also prevailing, can be learned, imitated and modified in time-series forecasting Transformer.\par
2. Currently, the outlier is tackled via DM self-attention, which roughly mask the entire diagonal of the self-attention score matrix, in FPPformer. Notice that the outliers shall be fewer than the normal segments of time-series sequences, which implies that the majority of masked patches are indeed normal and the masking behavior can negatively influences the feature extraction of input sequences. We believe that applying a prior anomaly detection method to each input sequence before forecasting and then only masking the detected anomalous patches can be a better format of utilizing the DM self-attention.\par
Both of the foregoing two limits and potential solutions will be our future research directions.\par}
\section{Conclusion}
In this work, we attempt to further develop the time-series forecasting Transformer from the perspective of decoder. We lucubrate the existing decoder designs, point out their drawbacks and propose our solutions. The ultimate product, i.e., FPPformer, achieves state-of-the-art performaces in multiple benchmarks, including multivariate and univariate ones, leveraging from refined attention mechanism and enhanced encoder-decoder architecture proposed by us.\par
\section*{Acknowledgment}
This work was partially supported by National Natural Science Foundation of China under grant \#U19B2033 and No.62103414. \par

\bibliographystyle{IEEEtran}
\bibliography{reference.bib}

\vfill

\end{document}